%% file: main.tex
\documentclass[11pt]{article}

\usepackage[final]{acl}

\usepackage{times}
\usepackage{latexsym}

\usepackage[T1]{fontenc}

\usepackage[utf8]{inputenc}

\usepackage{microtype}

\usepackage{inconsolata}

\usepackage{graphicx}

\usepackage{booktabs}
\usepackage{amssymb}
\usepackage{amsmath}
\usepackage[table]{xcolor} 
\usepackage{lipsum}
\usepackage{xcolor}   
\usepackage{pifont}
\usepackage{tabularx}
\usepackage{fontawesome}
\usepackage{longtable}
\usepackage{float}
\usepackage{placeins}
\usepackage{marvosym}

\newcommand{\cmark}{\textcolor{green!80!black}{\ding{51}}} 
\newcommand{\xmark}{\textcolor{red}{\ding{55}}} 

\newcommand\blfootnote[1]{%
  \begingroup
  \renewcommand\thefootnote{}\footnote{#1}%
  \addtocounter{footnote}{-1}%
  \endgroup
}

\title{Scaling Beyond Context: A Survey of Multimodal Retrieval-Augmented Generation for Document Understanding}

\author{
 \textbf{Sensen Gao\textsuperscript{1*}},
 \textbf{Shanshan Zhao\textsuperscript{2\Letter}},
 \textbf{Xu Jiang\textsuperscript{3}},
 \textbf{Lunhao Duan\textsuperscript{4*}},
 \textbf{Yong Xien Chng\textsuperscript{3*}},
 \\
 \textbf{Qing-Guo Chen\textsuperscript{2}},
 \textbf{Weihua Luo\textsuperscript{2}},
 \textbf{Kaifu Zhang\textsuperscript{2}},
 \textbf{Jia-Wang Bian\textsuperscript{5}},
 \textbf{Mingming Gong\textsuperscript{1,6\Letter}}
\\
\\
 \textsuperscript{1}MBZUAI,
 \textsuperscript{2}Alibaba Group,
 \textsuperscript{3}Tsinghua University,
 \\
 \textsuperscript{4}Wuhan University,
 \textsuperscript{5}Nanyang Technological University,
 \textsuperscript{6}University of Melbourne
\\
}

\begin{document}
\maketitle

\blfootnote{* This work was done during an internship at Alibaba Group.}
\blfootnote{\Letter Correspondence: \href{mailto:sshan.zhao00@gmail.com}{sshan.zhao00@gmail.com} and
   \href{mailto:mingming.gong@unimelb.edu.au}{mingming.gong@unimelb.edu.au}.}

\begin{abstract}
Document understanding is critical for applications from financial analysis to scientific discovery. Current approaches, whether OCR-based pipelines feeding Large Language Models (LLMs) or native Multimodal LLMs (MLLMs), face key limitations: the former loses structural detail, while the latter struggles with context modeling. Retrieval-Augmented Generation (RAG) helps ground models in external data, but documents’ multimodal nature, \emph{i.e.}, combining text, tables, charts, and layout, demands a more advanced paradigm: Multimodal RAG. This approach enables holistic retrieval and reasoning across all modalities, unlocking comprehensive document intelligence. Recognizing its importance, this paper presents a systematic survey of Multimodal RAG for document understanding. We propose a taxonomy based on domain, retrieval modality, and granularity, and review advances involving graph structures and agentic frameworks. We also summarize key datasets, benchmarks, applications and industry deployment, and highlight open challenges in efficiency, fine-grained representation, and robustness, providing a roadmap for future progress in document AI\footnote{Project is available at: \url{https://github.com/SensenGao/Multimodal-RAG-Survey-For-Document}.}.
\end{abstract}

\input{section/Introduction}
\input{section/Preliminary}

\input{section/Method}
\input{section/DB}
\input{section/Application}
\input{section/Conclusion}
\input{section/Limitation}

\section*{Acknowledgments}
This research project was supported by ARC grant DP240102088 and WIS-MBZUAI grant 142571.

\bibliography{main}

\appendix
\newpage
\label{sec:appendix}
In the appendix, a more detailed introduction to datasets and benchmarks is first provided, together with comprehensive evaluations of representative methods on these benchmarks (Appendix~\ref{sec:appendix_db}). Appendix~\ref{sec:appendix_eval} then presents evaluation metrics for multimodal RAG, explicitly distinguishing retrieval-oriented and generation-oriented assessments, followed by a systematic overview of commonly used training loss functions and interpretations of their roles in multimodal RAG systems (Appendix~\ref{sec:appendix_loss}). Beyond methodological foundations, open challenges and future research directions of multimodal RAG systems are extensively discussed in Appendix~\ref{sec:challenge}, while a focused critical analysis examining fundamental limitations, unresolved tensions, and representative failure cases is presented in Appendix~\ref{sec:critical_analysis}. Practical considerations for industrial deployment and real-world usage are analyzed in Appendix~\ref{sec:industrial-deployment}. In addition, the integration of multimodal RAG with agent-based and graph-based paradigms is examined in greater depth, with detailed analyses provided in Appendix~\ref{appendix:graph} and~\ref{appendix:agent}, respectively. Finally, Appendix~\ref{sec:appendix_key_contribution} summarizes the key contributions of all reviewed methods, offering a concise reference for rapidly understanding their core ideas.

\input{appendix/datasets}

\input{appendix/eval}

\input{appendix/loss}

\input{section/Challenge}
\input{appendix/hybrid}
\input{table/dataset}
\input{table/key_contribution}
\input{appendix/key_contribution}

\end{document}

%% file: section/Introduction.tex
\section{Introduction}
\begin{figure}[h]
    \centering
    \includegraphics[width=0.9\linewidth]{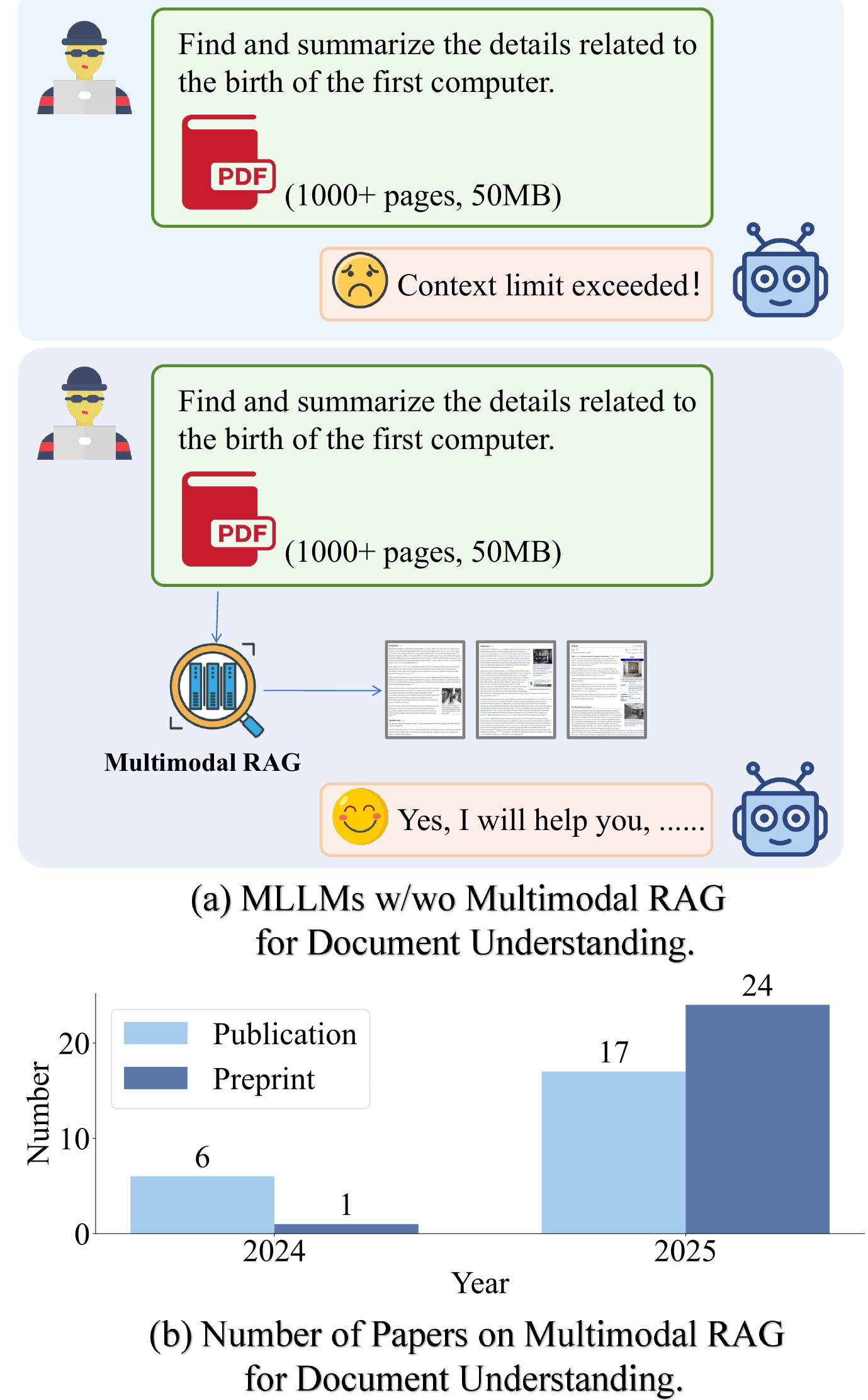}
    \vspace{-3mm}
    \caption{\textbf{Impact and research progress of Multimodal RAG for document understanding:} (a) MLLMs with and without Multimodal RAG for large document comprehension. (b) Growth in related publications from 2024 to 2025.}
    \label{fig:number_papers}
    \vspace{-6mm}
\end{figure}

Document understanding has become a pivotal task in the era of information explosion, as it empowers machines to automatically interpret, organize, and reason over the massive volumes of unstructured and semi-structured documents produced across diverse domains~\citep{subramani2020survey,ding2024deep}. Early studies primarily focus on text-centric documents, relying on optical character recognition (OCR) techniques~\citep{gu2021unidoc,appalaraju2021docformer,shi2016end} to support layout analysis and key information extraction. However, in real-world scenarios, particularly in scientific domains, documents are often visually rich and contain complex elements such as tables, charts, and images~\citep{park2019cord,ding2025vrd}. With the rapid progress of Large Language
Models (LLMs) and the rising demand for understanding increasingly complex and diverse document types, developing robust and generalizable document understanding frameworks has become an area of growing interest.


\input{table/benchmark_tokens}

In visually rich document understanding, different approaches have emerged to address the challenges of integrating layout, text, and structural information. Multimodal LLM (MLLM)-native methods commonly represent documents as long image sequences, enabling unified learning across modalities with MLLMs~\citep{duan2025docopilot,xiong2025docr1,yu2025docthinker,zhou2024doge,nassar2025smoldocling,ye2023mplug,hu2024mplug,hu2024mplug2}. While effective, these models struggle with very long documents spanning hundreds or thousands of pages, where sequence length limitations can hinder accurate retrieval and increase the risk of hallucination~\citep{deng2024longdocurl,ma2024mmlongbench}. As shown in Table~\ref{tab:bench_tokens}, current multimodal RAG benchmarks require 20–200M visual tokens, far exceeding the typical 128K–1M context limits of existing MLLMs~\citep{yang2025qwen3,achiam2023gpt,team2023gemini}.
To improve modularity and robustness, agent-based approaches introduce specialized agents for subtasks such as layout analysis, content extraction, instruction decomposition, and verification~\citep{liu2025hm,han2025mdocagent,wang2025multi,wu2150tabagent,yu2025visual}, though such designs often increase system complexity due to coordination overhead. Retrieval-augmented generation (RAG) methods provide another direction by grounding responses with external knowledge, typically retrieving the top-K most relevant pages (see Figure~\ref{fig:number_papers} (a)) across one or more documents~\citep{lewis2020retrieval}. Importantly, these paradigms are not mutually exclusive: RAG-based systems may employ agents to manage retrieval and verification, while agent-based workflows often incorporate RAG as one of the agent nodes, yielding more flexible hybrid frameworks. These complementary perspectives have shaped the landscape of document understanding, yet among them, RAG has drawn particular attention for its practicality and rapid growth~\citep{arslan2024survey,fan2024survey}.

Early RAG studies mainly rely on text-centric strategies, extracting text via OCR or combining OCR with MLLM-generated captions for visually rich documents, followed by encoding for retrieval~\citep{wang2022text,li2023towards,chen2024bge,khattab2020colbert}. Despite their effectiveness in certain scenarios, such text-based approaches exhibit fundamental limitations in handling visually rich documents, as they fail to adequately capture cross-modal cues and structural semantics~\citep{abootorabi2025ask,mei2025survey}. To address these shortcomings, recent efforts have increasingly focused on multimodal RAG frameworks. The growth trend in the number of papers is shown in Figure~\ref{fig:number_papers} (b). These methods often represent multi-page documents as image sequences~\citep{faysse2024colpali,yu2024visrag}, enabling visual encoders to extract richer representations for retrieval. Recent advances in multimodal RAG have increasingly emphasized finer-grained modeling within individual pages, including tables, charts, and other structured elements, to improve retrieval accuracy and robustness~\citep{wang2025vrag,choi2025zero}. Extending beyond these coarse-to-fine refinements, recent studies have also investigated graph-based indexing~\cite{yuan2025mkg} and multi-agent frameworks~\citep{liu2025hm}, which provide complementary mechanisms for structured reasoning and collaborative coordination in multimodal RAG.

This rapid evolution and increasing complexity in the field have naturally prompted efforts to synthesize the existing literature. However, a closer look reveals a significant gap. Prior surveys have reviewed RAG from multiple perspectives~\citep{arslan2024survey,fan2024survey,gao2023retrieval,hu2024rag,gupta2024comprehensive,zhao2024retrieval,church2024emerging}. In parallel, recent surveys examining multimodal RAG~\citep{zhao2023retrieving,abootorabi2025ask,mei2025survey} offer limited coverage of document understanding, typically discussing only a few relevant methods. Conversely, while document understanding has been extensively reviewed~\citep{subramani2020survey,ding2024deep,nandi2024visual,van2023document,ding2025survey}, existing surveys rarely address multimodal RAG.
To bridge this gap, we present the first comprehensive survey that explicitly connects multimodal RAG and document understanding. Unlike prior works that emphasize one aspect while overlooking the other, our survey systematically analyzes their intersection and organizes the most extensive collection of studies in this emerging field. Our contributions can be summarized as follows: (1) We present a comprehensive survey that categorizes existing methods by domain, retrieval modality, granularity, and hybrid enhancements, offering a structured perspective for future research. (2) We compile a broad collection of multimodal RAG datasets, benchmarks, and comparative results for systematic evaluation, and survey evaluation metrics spanning both retrieval and generation.
Together, these contributions outline a coherent landscape of multimodal RAG for document understanding, providing both a reference and guidance for future progress.


%% file: table/benchmark_tokens.tex
\begin{table}[t]
\centering
\setlength{\tabcolsep}{6pt}
\resizebox{\linewidth}{!}{
\begin{tabular}{l l r r}
\hline
Benchmark & Scope & \# Pages & Visual Tokens \\
\hline
M3DocVQA~\citep{cho2024m3docrag}     & Open-Domain & $\sim$40K  & $\sim$41M  \\
VisDoMBench~\citep{suri2025visdom} & Open-Domain & $\sim$21K   & $\sim$21M  \\
OpenDocVQA~\citep{tanaka2025vdocrag}  & Open-Domain & $\sim$206K & $\sim$206M \\
\hline
\end{tabular}
}
\vspace{-2mm}
\caption{Scale of representative document RAG benchmarks. Visual tokens are estimated assuming $\sim$1K visual tokens per page.}
\label{tab:bench_tokens}
\vspace{-6mm}
\end{table}

%% file: section/Preliminary.tex
\section{Preliminary}
In RAG, a system retrieves a set of relevant document pages and then generates a response conditioned on that evidence. Retrieval can be \emph{closed-domain} (\emph{e.g.}, grounding to a single source document) or \emph{open-domain} (searching a large corpus). We denote the candidate pool by $D=\{d_i\}_{i=1}^{N}$. Each $d_i$ may include a raster image as well as OCR text $T_i$. Using modality-specific encoders, we map queries and documents into a shared embedding space. Our notation uses lower-case symbols with subscripts for vectors (\emph{e.g.}, $z_i, e_q$), and we compute similarity using inner products.  Typically, the query $q$ is text, so we compute both text–text and text–image similarities in this shared space (and, if $q$ includes images, $e_q^{\mathrm{img}}$ can be defined analogously).

To embed documents and queries, we use image and text encoders: $z_i^{\mathrm{img}}=\mathrm{Enc}_{\mathrm{img}}(d_i)$, $z_i^{\mathrm{text}}=\mathrm{Enc}_{\mathrm{text}}(T_i)$, and $e_q^{\mathrm{text}}=\mathrm{Enc}_{\mathrm{text}}(q)$. Within each modality pair, similarities are inner products (optionally with unit-norm embeddings so the score is cosine similarity): $s_{\mathrm{text}}(e_q,z_i)=\langle e_q^{\mathrm{text}}, z_i^{\mathrm{text}}\rangle$ and $s_{\mathrm{img}}(e_q,z_i)=\langle e_q^{\mathrm{text}}, z_i^{\mathrm{img}}\rangle$.

\paragraph{Vision-only retrieval.}
When using only the image channel (\emph{i.e.}, for text-image similarity), we rank documents with the score $s_{\mathrm{img}}(e_q,z_i)$ and select those that exceed a threshold $\tau_{\mathrm{img}}$ (or simply take the $K$ results):
\begin{equation}
X_{\mathrm{img}}
= \left\{\, d_i \in D \;\middle|\; s_{\mathrm{img}}(e_q,z_i) \ge \tau_{\mathrm{img}} \,\right\}.
\end{equation}

\paragraph{Joint vision–text retrieval.}
We consider two widely used strategies.

\textbf{(a) Confidence-weighted score fusion.}
Image and text scores are combined with a convex weight that reflects per-item or per-query confidence. Let $\lambda_i\in[0,1]$ denote the image confidence for $d_i$ (\emph{e.g.}, from calibration or OCR quality); setting $\lambda_i{=}1$ recovers vision-only and $\lambda_i{=}0$ text-only:
\begin{equation}
\begin{aligned}
s_{\mathrm{conf}}(e_q,z_i)
  &= \lambda_i\, s_{\mathrm{img}}(e_q,z_i) \\
   &\quad + \bigl(1-\lambda_i\bigr)\, s_{\mathrm{text}}(e_q,z_i), \\
X_{\mathrm{conf}}
  &= \left\{\, d_i \in D \;\middle|\; s_{\mathrm{conf}}(e_q,z_i) \ge \tau_{\mathrm{conf}} \,\right\}.
\end{aligned}
\end{equation}

\textbf{(b) Union of modality-specific pages.}
This strategy involves retrieving pages with each modality independently and then forming the union of the results (optionally followed by deduplication or rank fusion such as Borda or reciprocal-rank fusion~\citep{cormack2009reciprocal,calumby2017rank}) using modality-aware thresholds $\tau_{\mathrm{img}},\tau_{\mathrm{text}}$:
\begin{equation}
\begin{aligned}
X_{\mathrm{img}}  &= \left\{\, d_i \in D \;\middle|\; s_{\mathrm{img}}(e_q,z_i)  \ge \tau_{\mathrm{img}}  \,\right\},\\
X_{\mathrm{text}} &= \left\{\, d_i \in D \;\middle|\; s_{\mathrm{text}}(e_q,z_i) \ge \tau_{\mathrm{text}} \,\right\},\\
X_{\cup}          &= X_{\mathrm{img}} \cup X_{\mathrm{text}}.
\end{aligned}
\end{equation}
(Equivalently, one may use top-$K$ per modality and take the union $X_{\cup}^{(K)}$.)

\paragraph{Generation.}
A generator $\mathcal{G}$ conditions on the original query and the retrieved context chosen as $X_{\mathrm{img}}$, $X_{\mathrm{conf}}$, or $X_{\cup}$ depending on the retrieval regime and produces the final response. The specific mechanism for aggregating multiple pages (\emph{e.g.}, via cross-attention or learned pooling) is left abstract:
\begin{equation}
r = \mathcal{G}(q, X).
\end{equation}

\input{table/test}

%% file: table/test.tex
\begin{table*}[ht]
\centering
\setlength{\tabcolsep}{4pt}
\resizebox{\linewidth}{!}{
\begin{tabular}{lccccccccccc}
\toprule
\textbf{Method} & 
\textbf{Venue} & 
\textbf{LLM/VLM} & 
\textbf{Vision Encoder} &
\textbf{Training} & 
\textbf{OCR} & 
\textbf{Domain} & 
\textbf{Modality} & 
\textbf{Granularity} &
\textbf{Graph} &
\textbf{Agent} \\
\midrule
DSE~\citeyearpar{ma2024unifying} & EMNLP & Phi3V & CLIP-ViT-L/14 & \cmark & \xmark & Open & Image & Page & \xmark & \xmark \\
ColPali~\citeyearpar{faysse2024colpali} & ICLR & PaliGemma-3B & SigLIP-SO400M & \cmark & \xmark & Open & Image & Page & \xmark & \xmark \\
ColQwen2~\citeyearpar{faysse2024colpali} & ICLR & Qwen2-VL-2B & ViT-BigG & \cmark & \xmark & Open & Image & Page & \xmark & \xmark \\  
CREAM~\citeyearpar{zhang2024cream} & ACM MM & LLaMA2-7B & Pix2Struct & \cmark & \cmark & Closed & Image+Text & Page & \xmark & \xmark \\
VisRAG~\citeyearpar{yu2024visrag} & ICLR & MiniCPM-V2.0 & SigLIP-SO400M & \cmark & \xmark & Open & Image & Page & \xmark & \xmark \\
SV-RAG~\citeyearpar{chen2024sv} & ICLR & InternVL2-4B & InternViT-300M & \cmark & \xmark & Closed & Image & Page & \xmark & \xmark \\
M3DocRAG~\citeyearpar{cho2024m3docrag} & Preprint & Qwen2-VL-7B &  ViT-BigG & \xmark & \xmark & Open & Image & Page & \xmark & \xmark \\
VisDoMRAG~\citeyearpar{suri2025visdom} & NAACL & Qwen2-VL-2B & ViT-BigG & \xmark & \cmark & Open & Image+Text & Page & \xmark & \xmark \\
GME~\citeyearpar{zhang2025bridging} & CVPR & Qwen2-VL-7B & ViT-BigG & \cmark & \cmark & Open & Image+Text & Page & \xmark & \xmark \\
ViDoRAG~\citeyearpar{wang2025vidorag} & EMNLP & Qwen2.5-VL-7B & ViT-BigG & \xmark & \cmark & Open & Image+Text & Page & \xmark & \cmark \\
HM-RAG~\citeyearpar{liu2025hm} & ACM MM & Qwen2.5-VL-7B & ViT-BigG & \xmark & \cmark & Open & Image+Text & Page & \cmark & \cmark \\
VDocRAG~\citeyearpar{tanaka2025vdocrag} & CVPR & Phi3V &  CLIP-ViT-L/14 & \cmark & \xmark & Open & Image & Page & \xmark & \xmark \\
FRAG~\citeyearpar{huang2025frag} & Preprint & InternVL2-8B & InternViT-300M & \xmark & \xmark & Closed & Image & Page & \xmark & \xmark \\
MG-RAG~\citeyearpar{xu2025multi} & Preprint & Qwen2.5-VL-3B-Instruct & ViT-BigG & \xmark & \cmark & Closed & Image+Text & Element & \xmark & \xmark \\
VRAG-RL~\citeyearpar{wang2025vrag} & Preprint & Qwen2.5-VL-7B-Instruct & ViT-BigG & \cmark & \xmark & Open & Image & Element & \xmark & \xmark \\
CoRe-MMRAG~\citeyearpar{tian2025core} & ACL & Qwen2-VL-7B & ViT-BigG & \cmark & \cmark & Open & Image+Text & Page & \xmark & \xmark \\
Light-ColPali~\citeyearpar{ma2025towards} & ACL & PaliGemma & SigLIP-SO400M & \cmark & \xmark & Open & Image & Page & \xmark & \xmark \\
MM-R5~\citeyearpar{xu2025mm} & Preprint & Qwen2.5-VL-7B & ViT-BigG & \cmark & \xmark & Open & Image & Page & \xmark & \xmark \\
SimpleDoc~\citeyearpar{jain2025simpledoc} & Preprint & Qwen2.5-VL-3B-Instruct & ViT-BigG & \xmark & \xmark & Open & Image+Text & Page & \xmark & \xmark \\
VisChunk~\citeyearpar{tripathi2025vision} & Preprint & Gemini-2.5-Pro & -- & \xmark & \cmark & Closed & Image+Text & Page & \xmark & \xmark \\
DocVQA-RAP~\citeyearpar{yu2025beyond} & ICIC & Qwen2-VL-2B & ViT-BigG & \xmark & \xmark & Open & Image & Element & \xmark & \xmark \\
RL-QR~\citeyearpar{cha2025generalized} & Preprint & Qwen2.5-VL-3B-Instruct &  ViT-BigG & \cmark & \xmark & Open & Image & Page & \xmark & \xmark \\
MMRAG-DocQA~\citeyearpar{gong2025mmrag} & Preprint & Qwen-VL-Plus & ViT-BigG & \xmark & \cmark & Closed & Image+Text & Element & \xmark & \xmark \\
Patho-AgenticRAG~\citeyearpar{zhang2025patho} & Preprint & Qwen2-VL-2B & ViT-BigG & \cmark & \xmark & Open & Image & Page & \xmark & \cmark \\
M2IO-R1~\citeyearpar{xiao2025m2io} & Preprint & BGE-M3 & -- & \cmark & \cmark & Open & Image+Text & Page & \xmark & \xmark \\
mKG-RAG~\citeyearpar{yuan2025mkg} & Preprint &  LLaMA-3.1-8B & CLIP ViT-L/14 & \cmark & \cmark & Open & Image+Text & Element & \cmark & \xmark \\
DB3Team-RAG~\citeyearpar{xia2025db3} & Preprint & Llama 3.2-VL & CLIP ViT-L/14 & \cmark & \cmark & Open & Image+Text & Page & \cmark & \xmark \\
PREMIR~\citeyearpar{choi2025zero} & EMNLP & Qwen2.5-VL-72B & ViT-BigG & \xmark & \cmark & Open & Image+Text & Element & \xmark & \xmark \\
ReDocRAG~\citeyearpar{lopez2025enhancing} & ICDAR WML & Qwen2.5-VL-7B-Instruct & ViT-BigG & \cmark & \xmark & Closed & Image & Page & \xmark & \xmark \\
CMRAG~\citeyearpar{chen2025cmrag} & Preprint & Qwen2.5-VL-7B-Instruct & ViT-BigG & \xmark & \cmark & Open & Image+Text & Page & \xmark & \xmark \\
MoLoRAG~\citeyearpar{wu2025molorag} & EMNLP & Qwen2.5-VL-7B & ViT-BigG & \cmark & \xmark & Open & Image & Page & \cmark & \xmark \\
SERVAL~\citeyearpar{nguyen2025serval} & Preprint & InternVL3-14B & InternViT-300M & \xmark & \xmark & Open & Image & Page & \xmark & \xmark \\
MetaEmbed~\citeyearpar{xiao2025metaembed} & Preprint & Qwen2.5-VL-32B & ViT-BigG & \cmark & \xmark & Open & Image & Page & \xmark & \xmark \\
DocPruner~\citeyearpar{yan2025docpruner} & Preprint & Qwen2.5-VL-3B-Instruct & ViT-BigG & \cmark & \xmark & Open & Image & Page & \xmark & \xmark \\
RECON~\citeyearpar{wangrecon} & Preprint & GPT-4o-mini & -- & \xmark & \xmark & Open & Image+Text & Element & \cmark & \xmark \\
LAD-RAG~\citeyearpar{sourati2025lad} & Preprint & GPT-4o-200b-128 & -- & \xmark & \xmark & Open & Image+Text & Element & \cmark & \xmark \\
HEAVEN~\citeyearpar{kim2025hybrid} & Preprint & Qwen2.5-VL-3B-Instruct & ViT-BigG & \xmark & \xmark & Open & Image & Page & \xmark & \xmark\\
DREAM~\citeyearpar{zhang2025dream} & ACM MM & InternVL2-40B & InternViT-6B & \cmark & \xmark & Closed & Image & Page & \xmark & \xmark \\
MARA~\citeyearpar{wu2025mara} & ACM MM & MiniCPM-V2.0 & SigLIP-SO400M & \cmark & \xmark & Open & Image & Element & \xmark & \xmark \\
HEAR~\citeyearpar{chen2025hear} & ACM MMW & Qwen2.5-VL-32B-Instruct & ViT-BigG & \xmark & \cmark & Closed & Image+Text & Page & \xmark & \cmark \\
HPC-ColPali~\citeyearpar{bach2025hierarchical} & Preprint & PaliGemma-3B & SigLIP-SO400M & \cmark & \xmark & Open & Image & Page & \xmark & \xmark \\
RegionRAG~\citeyearpar{li2025regionrag} & Preprint & Qwen2.5-VL-3B & ViT-BigG & \cmark & \xmark & Open & Image & Element & \xmark & \xmark \\
IndustryRAG~\citeyearpar{lim2025distilling} & EMNLP Industry & Qwen2.5-VL-32B-Instruct & ViT-BigG & \xmark & \cmark & Open & Image & Page & \xmark & \xmark \\
COLMATE~\citeyearpar{masry2025colmate} & EMNLP Industry & PaliGemma-3B & SigLIP-SO400M & \cmark & \xmark & Open & Image & Page & \xmark & \xmark \\
LILaC~\citeyearpar{yun2025lilac} & EMNLP & MM-Embed & -- & \xmark & \xmark & Open & Image & Element & \cmark & \xmark \\
HKRAG~\citeyearpar{tong2025hkrag} & Preprint & Phi3V &  CLIP-ViT-L/14 & \cmark & \xmark & Open & Image & Element & \xmark & \xmark \\
SLEUTH~\citeyearpar{liu2025resolving} & Preprint & PaliGemma-3B & SigLIP-SO400M & \xmark & \xmark & Open & Image & Page & \xmark & \cmark \\
Snappy~\citeyearpar{georgiou2025spatially} & Preprint & PaliGemma-3B & SigLIP-SO400M & \xmark & \cmark & Open & Image & Element & \xmark & \xmark \\
\bottomrule
\end{tabular}
}
\vspace{-2mm}
\caption{\textbf{Comparison of recent Multimodal RAG methods for document understanding.} The table summarizes methods along the following dimensions: venue, backbone LLM/VLM, vision encoder, training status, OCR integration, domain scope, retrieval modality, retrieval granularity, graph incorporation, and agent usage.}
\label{tab:all_methods}
\vspace{-2mm}
\end{table*}

%% file: section/Method.tex
\section{Key Innovations and Methodologies}
\label{sec:innovations}
\begin{figure}[t]
    \centering
    \includegraphics[width=0.95\linewidth]{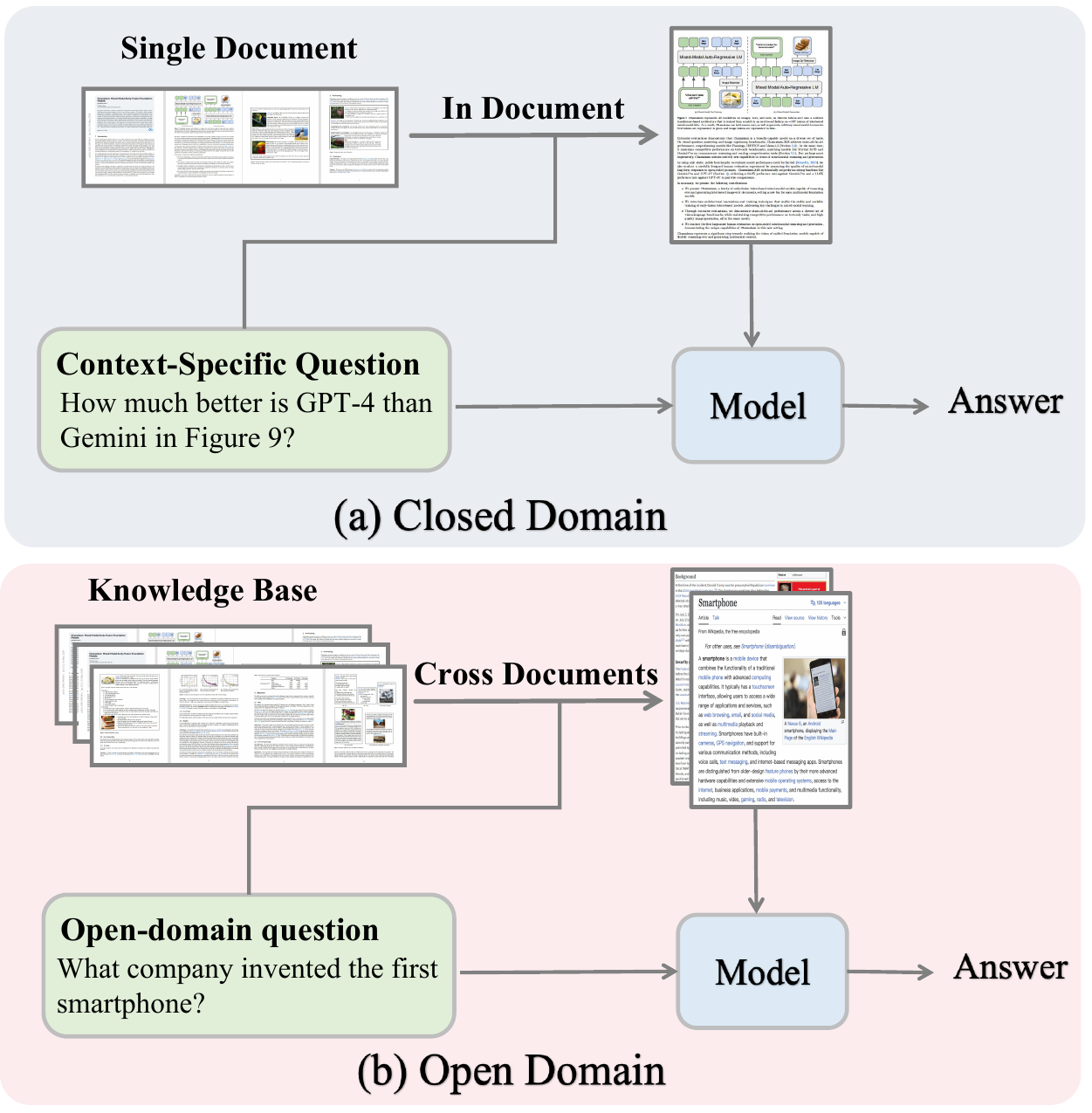}
    \caption{\textbf{Comparison between closed-domain and open-domain multimodal RAG}. (a) In the closed domain, the model leverages in-document retrieval from a single document to answer context-specific questions. (b) In the open domain, the model relies on cross-document retrieval from multiple documents to answer open-ended questions.}
    \label{fig:domain}
    \vspace{-4mm}
\end{figure}

In this section, we examine the core innovations and methodological advances in recent multimodal RAG approaches for document understanding. Table~\ref{tab:all_methods} presents a systematic comparison of representative methods along several key dimensions, including domain openness, retrieval modality, retrieval granularity, graph-based integration, and agent-based enhancement. To provide a structured discussion, we elaborate on each dimension in turn: the distinction between open- and closed-domain settings (Section~\ref{subsec:domain}), the impact of retrieval modality (Section~\ref{subsec:modality}), the role of retrieval granularity (Section~\ref{subsec:granularity}), agent and graph based hybrid enhancements (Section~\ref{subsec:hybrid_enhance}). 

\subsection{Open and Closed Domain}
\label{subsec:domain}
RAG addresses the limitations of LLMs in knowledge acquisition, such as knowledge cut-off, and extends their applicability to specialized domains~\citep{lewis2020retrieval,jorensufficient,ye2024r2ag,gupta2024comprehensive,huang2024survey,cheng2025survey}. For document understanding, open-domain multimodal RAG retrieves information from large corpora of domain-specific documents to construct extensive knowledge bases. In contrast, closed-domain multimodal RAG focuses on a single document and selects only the most relevant pages for retrieval, thereby reducing input length and mitigating issues related to limited context windows and hallucination. The distinction between open-domain and closed-domain multimodal RAG is illustrated in Figure~\ref{fig:domain}.

\textbf{Open-Domain Multimodal RAG.}
Open-domain multimodal RAG enhances an LLM’s knowledge in specialized domains by constructing retrieval databases from large collections of documents. Early approaches typically apply OCR to all documents to build text-based retrieval indices~\citep{wang2022text,li2023towards,chen2024bge,khattab2020colbert}, but this process is computationally expensive and inefficient. To improve scalability, recent methods such as DSE~\citep{ma2024unifying} and ColPali~\citep{faysse2024colpali} leverage vision-language models (VLMs) to encode document pages directly into image embeddings, achieving significant efficiency gains. Despite these advances, most approaches still focus on reasoning within single documents and lack explicit mechanisms for integrating knowledge across sources. Addressing this limitation, M3DocRAG~\citep{cho2024m3docrag} introduces approximate indexing to accelerate large-scale retrieval and establishes the benchmark M3DocVQA with over 3,000 documents, while VDocRAG~\citep{tanaka2025vdocrag} constructs the OpenDocVQA dataset and mitigates page-level information loss by compressing visual content into dense token representations aligned with text.

\begin{figure}[t]
    \centering
    \includegraphics[width=0.95\linewidth]{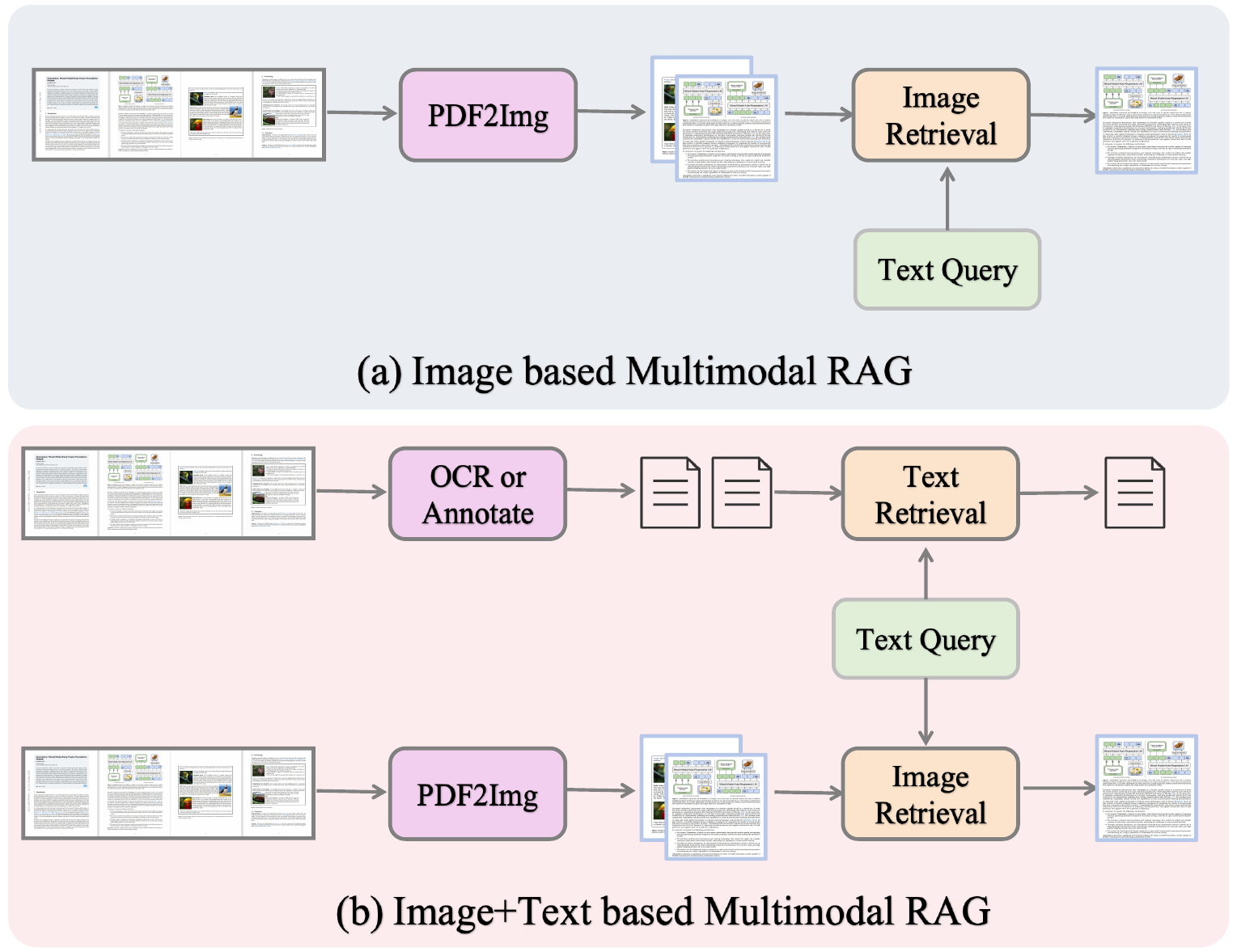}
    \caption{\textbf{Comparison of retrieval modality:} (a) image-based RAG retrieves information solely from page images, offering efficiency but limited textual detail; (b) image+text-based RAG integrates OCR/annotations with visual features, enabling richer retrieval at the cost of higher processing complexity.}
    \label{fig:mode}
    \vspace{-3mm}
\end{figure}

\textbf{Closed-Domain Multimodal RAG.}
Closed-domain multimodal RAG is designed for practical scenarios where MLLMs encounter difficulties with extremely long documents or videos. Current MLLMs remain constrained by limited context windows, and long-context processing often amplifies the risk of hallucination. To address this, closed-domain approaches retrieve only the most relevant segments (\emph{e.g.}, pages or frames) from a target document and provide them as input to the MLLM, thereby improving both efficiency and reliability. For single-document visual question answering (DocVQA), SV-RAG~\citep{chen2024sv} employs the MLLM itself as a multimodal retriever, with specialized adapters for page retrieval and evidence-based reasoning. FRAG~\citep{huang2025frag}, by contrast, independently scores each frame or page, applies a Top-K selection to retain the most informative content, and then delegates answer generation to existing LMMs. CREAM~\citep{zhang2024cream} introduces a coarse-to-fine multimodal retrieval and attention-pooling integration framework, enabling effective cross-page reasoning and multi-page document comprehension for visual question answering. All approaches demonstrate that closed-domain multimodal RAG enables effective comprehension of long documents and videos without extending the model’s context length.

\subsection{Retrieval Modality}
\label{subsec:modality}
Early text-only RAG methods rely exclusively on textual signals for retrieval, which limits their practical utility: they require time-consuming OCR and underperform on visually rich documents. To address these limitations, current research advances multimodal RAG. One approach treats each page as an image and encodes it with the vision encoder of a VLM. Another adopts hybrid designs that pair page-level images with OCR-extracted text or auxiliary textual annotations generated by MLLMs. The resulting cross-modal representations then support retrieval independently or via score fusion, where similarity scores from different modalities combine to improve performance.

\textbf{Image-based Retrieval Modality.} To handle visually rich documents, most existing methods represent each page as an image and encode it with VLMs, using the VLMs' hidden states as page-level representations (see Figure~\ref{fig:mode} (a)). In parallel, the query is encoded, and page–query relevance is computed via similarity-based ranking~\citep{ma2024unifying,faysse2024colpali,yu2024visrag,chen2024sv,ma2025towards,yu2025beyond}. Building on image-based embeddings, MM-R5~\citep{xu2025mm} introduces a reasoning-enhanced reranker that combines supervised fine-tuning and reinforcement learning to strengthen instruction following, elicit explicit reasoning chains, and leverage task-specific rewards for greater precision and interpretability. Complementing this direction, Light-ColPali~\citep{ma2025towards} and HPC-ColPali~\citep{bach2025hierarchical} improve the efficiency of ColPali-style multi-vector retrieval by compressing patch-level representations, reducing memory and computation while largely preserving retrieval accuracy.

\textbf{Image+Text based Retrieval Modality.} Leveraging both image and text for retrieval mitigates the loss of fine-grained textual cues that arise when relying solely on page-level VLM encoders. The text channel is derived from OCR~\citep{zhang2024cream,suri2025visdom,liu2025hm,wang2025vidorag} or from summary annotations generated by large VLMs~\citep{jain2025simpledoc,choi2025zero} (see Figure~\ref{fig:domain} (b)). VisDoMRAG~\citep{suri2025visdom} and HM-RAG~\citep{liu2025hm} adopt a dual-path pipeline: they retrieve and reason within each modality, then summarize and fuse the results into a single answer. By contrast, ViDoRAG~\citep{wang2025vidorag} and PREMIR~\citep{choi2025zero} also retrieve per modality but merge candidates via a simple union before answer generation. Complementing these designs, SimpleDoc~\citep{jain2025simpledoc} uses a two-stage scheme for DocVQA: embedding-based candidate selection followed by re-ranking with VLM-generated page summaries, so that the summaries provide richer semantics for more precise evidence aggregation.

\begin{figure}[t]
    \centering
    \includegraphics[width=0.95\linewidth]{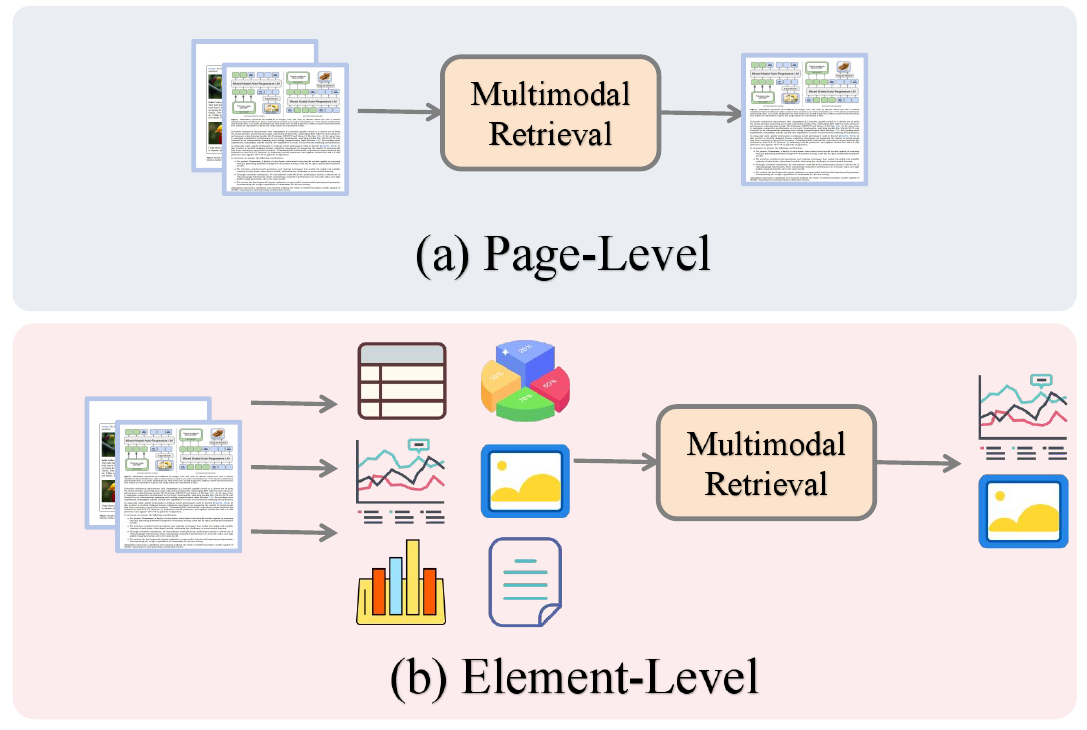}
    \caption{\textbf{Comparison of retrieval granularity in multimodal document search.} (a) Page-level: entire pages are encoded and ranked as whole units. (b) Element-level: pages are decomposed into tables, charts, images, and text blocks; retrieval operates over these elements to localize evidence and aggregate results.}
    \label{fig:granularity}
    \vspace{-3mm}
\end{figure}

\subsection{Retrieval Granularity}
\label{subsec:granularity}
In document-oriented multimodal RAG, early studies typically treat the page as the atomic retrieval unit, without modeling finer structures such as tables, charts, or layout cues (see Figure ~\ref{fig:granularity}). Recent work increasingly focuses on retrieval at a finer, within-page granularity. Some approaches explicitly encode these components to enhance retrieval accuracy, whereas others adopt a two-stage pipeline: first retrieve the most relevant pages, then perform retrieval within those pages to establish fine-grained grounding. This shift toward finer retrieval granularity enables models to deliver more precise and contextually grounded answers.

Recent multimodal RAG research demonstrates a clear evolution toward fine-grained, structure-aware evidence selection. VRAG-RL~\citep{wang2025vrag} leverages reinforcement learning for region guidance, while MG-RAG~\citep{xu2025multi} and MMRAG-DocQA~\citep{gong2025mmrag} enable multi-granularity retrieval via hierarchical indexing across pages and layouts. At the segment level, DocVQA-RAP~\citep{yu2025beyond} ranks segments to suppress redundancy. Beyond segmentation, mKG-RAG~\citep{yuan2025mkg} aligns cross-modal entities via knowledge graphs, whereas PREMIR~\citep{choi2025zero} matches queries against QA pairs for charts. Recent region-level methods like MARA~\citep{wu2025mara} and RegionRAG~\citep{li2025regionrag} introduce query-aligned representations and patch aggregation to reduce noise. Furthermore, HKRAG~\citep{tong2025hkrag} captures fine-print knowledge via hybrid masking, and Snappy~\citep{georgiou2025spatially} achieves efficient localization by propagating patch-level similarity. Collectively, these approaches illustrate the shift toward increasingly fine-grained retrieval in document-heavy systems.

\subsection{Hybrid Enhancements for Multimodal RAG}
\label{subsec:hybrid_enhance}
The main text focuses on integrating multimodal RAG with graph-based and agent-based methods. The Appendix~\ref{appendix:graph} and ~\ref{appendix:agent} extends this discussion to more advanced integrations, highlighting open challenges and future research directions.

\begin{figure}[t]
    \centering
    \includegraphics[width=0.95\linewidth]{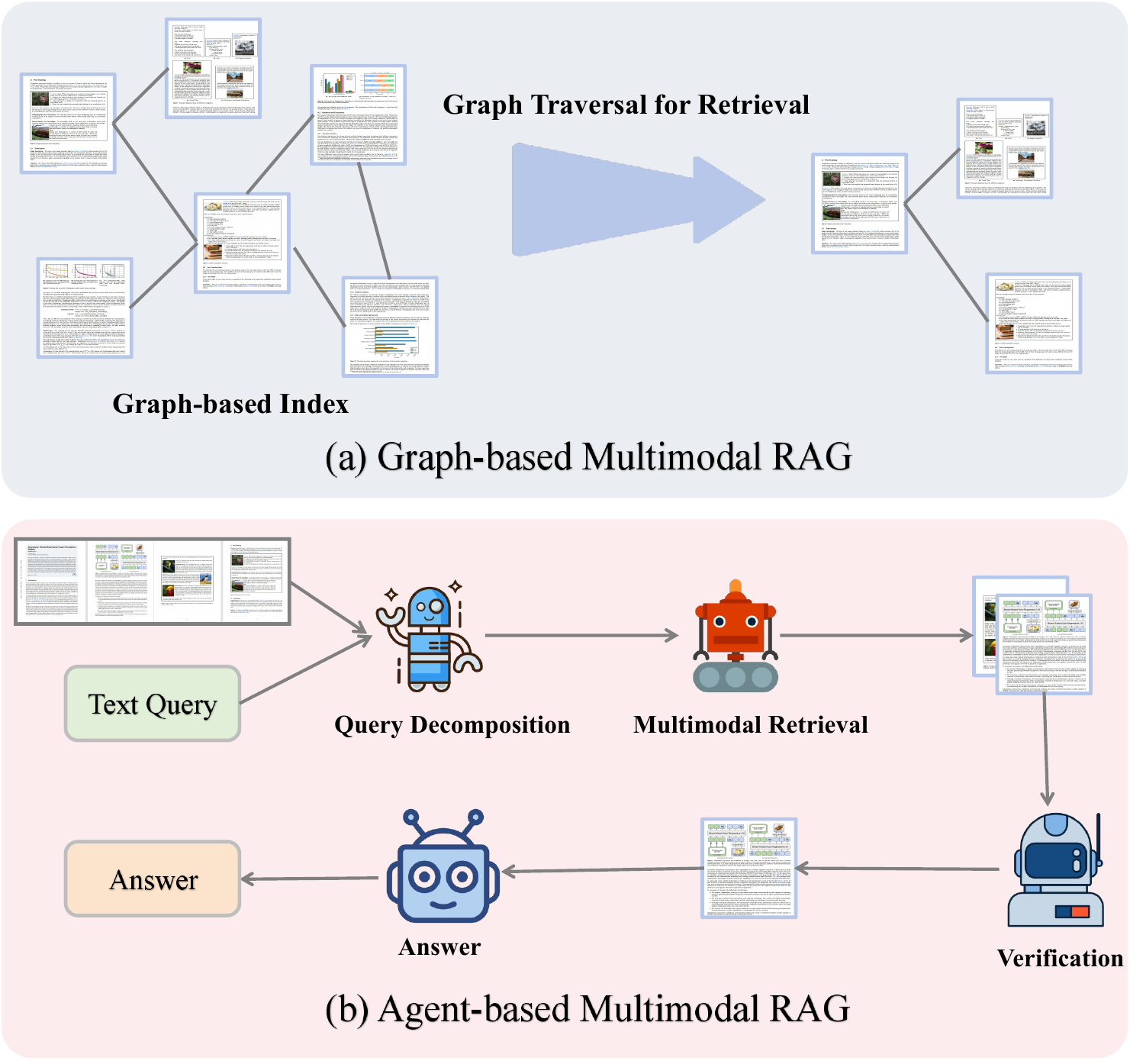}
    \caption{\textbf{Hybrid enhancements for multimodal RAG.} (a) Graph-based: documents/elements form a graph index, and retrieval proceeds via graph traversal to surface relevant neighborhoods. (b) Agent-based: an LLM agent decomposes the text query, orchestrates multimodal retrieval, verifies the gathered evidence, and synthesizes the final answer.}
    \label{fig:graph_agent}
    \vspace{-4mm}
\end{figure}

\paragraph{Graph-based Multimodal RAG.}
Graph-based multimodal RAG extends the framework by representing multimodal content as an explicit graph, as shown in Figure~\ref{fig:graph_agent} (a). Nodes denote modalities or atomic content units such as pages, text spans, images, tables, and layout blocks, while edges encode semantic, spatial, and contextual relations. Retrieval and reasoning over this multimodal graph integrate heterogeneous evidence more effectively, enable finer-grained grounding, and improve the robustness and interpretability of multimodal RAG systems.

HM-RAG~\citep{liu2025hm} introduces a hierarchical multi-agent framework utilizing graph databases to capture structured relations, while mKG-RAG~\citep{yuan2025mkg} explicitly constructs multimodal knowledge graphs to align entities across vision and text. Building on such structured repositories, DB3Team-RAG~\citep{xia2025db3} incorporates image-indexed graphs to handle complex ego-centric queries within domain-specific pipelines. Shifting focus to document topology, MoLoRAG~\citep{wu2025molorag} leverages page graphs to encode logical connections for multi-page understanding. This structure-aware modeling is further refined by RECON~\citep{wangrecon}, which builds a global graph linking intra-page visual relations with inter-page entity connections. Furthermore, LAD-RAG~\citep{sourati2025lad} and LILaC~\citep{yun2025lilac} focus on layout-aware component graphs, employing dynamic traversal or late interaction to support multi-hop reasoning. Collectively, these methods highlight the pivotal role of graph structures as either external repositories or internal document representations in advancing reliable multimodal retrieval.

\paragraph{Agent-based Multimodal RAG.}
Agent-based multimodal RAG extends this paradigm by employing autonomous agents to orchestrate retrieval–generation interactions across modalities. These agents dynamically formulate queries, select retrieval strategies, and adaptively fuse information from text, images, tables, and other modalities (see Figure~\ref{fig:graph_agent} (b)). Multi-agent collaboration further enables iterative reasoning, verification, and evidence refinement, improving the accuracy, reliability, and transparency of multimodal RAG systems.

ViDoRAG~\citep{wang2025vidorag} introduces an iterative agent workflow in which agents perform exploration, summarization, and reflection, improving multimodal retrieval and reasoning over visually rich documents. HM-RAG~\citep{liu2025hm} further extends this idea with a hierarchical multi-agent architecture, combining query decomposition, modality-specific parallel retrieval, and a decision agent that aggregates evidence through consistency voting and refinement. Adapting agentic RAG to the medical domain, Patho-AgenticRAG~\citep{zhang2025patho} enables task decomposition and multi-turn search to retrieve aligned text–image evidence from pathology textbooks while reducing diagnostic hallucinations. Along similar lines, HEAR~\citep{chen2025hear} and SLEUTH~\citep{liu2025resolving} focus on improving long-document understanding by coupling VLM-based parsing with closed-loop or coarse-to-fine agent reasoning, allowing cross-modal inconsistencies to be corrected and salient evidence to be distilled into compact contexts. Overall, these approaches demonstrate how diverse agent designs enhance fine-grained retrieval and reasoning in multimodal RAG systems.


%% file: section/DB.tex
\section{Dataset and Benchmark}
\input{table/datasets_overreview}

Datasets and benchmarks commonly used in multimodal RAG for document understanding typically consist of visually rich document collections. We compile the most widely adopted datasets and benchmarks for this task, reporting their query volume, dataset scale, and data types, such as text, tables, charts, and slides. The representative datasets and benchmarks are presented in the upper part of Table~\ref{tab:DB_overreview}. They support the training and evaluation of multimodal models and also serve as essential resources for constructing broader evaluation frameworks. Nevertheless, these resources still exhibit important limitations, motivating the development of more diverse and realistic benchmarks.

Many studies have revealed limitations in multimodal RAG systems, leading to the development of diverse benchmarks summarized in the lower half of Table~\ref{tab:DB_overreview}. ColPali~\citep{faysse2024colpali} proposes ViDoRe, a comprehensive benchmark covering academic and practical tasks across domains such as energy, government, and healthcare; while SV-RAG~\citep{chen2024sv} builds VISR-BENCH from a large-scale, manually validated dataset with high task diversity. To overcome single-document evaluation, M3DocVQA~\citep{cho2024m3docrag}, VisDoMRAG~\citep{suri2025visdom}, and VDocRAG~\citep{tanaka2025vdocrag} extend evaluation to cross-document open-domain scenarios using M3DocVQA, VisDoMBench, and OpenDocVQA, respectively. Focusing on large-scale retrieval closer to real-world applications, ViDoRAG~\citep{wang2025vidorag} introduces ViDoSeek, a benchmark for RAG evaluation with uniquely answerable queries. Furthermore, UniDoc-Bench~\citep{peng2025unidoc} establishes a document-centric MM-RAG benchmark, enabling systematic comparisons of multimodal retrieval and fusion strategies on real-world PDFs through unified text, table, and figure evidence linking. BBox-DocVQA~\citep{yu2025bbox} provides a DocVQA dataset with bounding-box grounding for supervision of spatial reasoning and evidence localization.

We also present the performance of different multimodal RAG methods across various benchmarks, along with a detailed explanation of the evaluation metrics and their computation. The full details are provided in Appendix~\ref{sec:appendix_db}.

%% file: table/datasets_overreview.tex
\newcommand{\CT}{\faFileTextO}        
\newcommand{\TB}{\faTable}            
\newcommand{\CH}{\faBarChart}         
\newcommand{\SL}{\faFilePowerpointO}  

\begin{table}[t]
\centering
\setlength{\tabcolsep}{4pt}
\resizebox{\linewidth}{!}{
\begin{tabular}{l c c c}
\toprule
\textbf{Dataset} & \textbf{\# Queries} & \textbf{\# Documents/Images} & \textbf{Content} \\
\midrule
TabFQuAD~\citeyearpar{d2020fquad} & 210 & 210\,(I) & \TB \\
PlotQA~\citeyearpar{methani2020plotqa} & 28.9M & 224K\,(I) & \CH \\
DocVQA~\citeyearpar{mathew2021docvqa} & 50K & 12,767\,(I) & \CT\,\TB\,\CH \\
VisualMRC~\citeyearpar{tanaka2021visualmrc} & 30,562 & 10,197\,(I) & \CT\,\TB\,\CH \\
TAT-DQA~\citeyearpar{zhu2022towards} & 16,558 & 2,758\,(D) & \CT\,\TB\,\CH \\
InfoVQA~\citeyearpar{mathew2022infographicvqa} & 30K & 5.4K\,(I) & \CT\,\TB\,\CH \\
ChartQA~\citeyearpar{masry2022chartqa} & 23.1K & 17.1K\,(I) & \CH \\
ScienceQA~\citeyearpar{saikh2022scienceqa} & 21K & 7,803\,(I) & \CT\,\TB\,\CH \\
DUDE~\citeyearpar{van2023document} & 41,491 & 4,974\,(D) & \CT\,\TB\,\CH \\
SlideVQA~\citeyearpar{tanaka2023slidevqa} & 52K & 14.5K\,(I) & \SL \\
ArXivQA~\citeyearpar{li2024multimodal} & 100K & 16.6K\,(D) & \CT\,\TB\,\CH \\
MMLongBench-Doc~\citeyearpar{ma2024mmlongbench} & 1,062 & 130\,(D) & \CT\,\TB\,\CH\,\SL \\
PaperTab~\citeyearpar{hui2024uda} & 393 & 307\,(D) & \CT\,\TB \\
FetaTab~\citeyearpar{hui2024uda} & 1,023 & 878\,(D) & \TB \\
SPIQA~\citeyearpar{pramanick2024spiqa} & 27K & 25.5K\,(D) & \TB\,\CH \\
LongDocUrl~\citeyearpar{deng2024longdocurl} & 2,325 & 396\,(D) & \CT\,\TB\,\CH \\
\midrule
ViDoRe~\citeyearpar{faysse2024colpali} & 3.8K & 8.3K\,(D) & \CT\,\TB\,\CH \\
VisR-Bench~\citeyearpar{chen2024sv} & 471 & 226\,(D) & \CT\,\TB\,\CH\,\SL \\
M3DoCVQA~\citeyearpar{cho2024m3docrag} & 2,441 & 3,368\,(D) & \CT\,\TB\,\CH \\
VisDoMBench~\citeyearpar{suri2025visdom} & 2,271 & 1,277\,(D) & \CT\,\TB\,\CH\,\SL \\
ViDoSeek~\citeyearpar{wang2025vidorag} & 1,142 & 300\,(D) & \CT\,\TB\,\CH \\
OpenDocVQA~\citeyearpar{tanaka2025vdocrag} & 206K & 43K\,(I) & \CT\,\TB\,\CH \\
UniDoc-Bench~\citeyearpar{peng2025unidoc} & 1.6K & 70K(I) & \CT\,\TB\,\CH \\
BBox-DocVQA~\citeyearpar{yu2025bbox} & 32K & 4.4K(D) & \CT\,\TB\,\CH \\
\bottomrule
\end{tabular}
}
\vspace{-2mm}
\caption{Overview of datasets and benchmarks in multimodal RAG for document understanding. 
We report the number of queries, dataset size, and covered content types (\CT{} Text, \TB{} Tables, \CH{} Charts, \SL{} Slides). 
(D) and (I) indicate that the count refers to documents or images, respectively. 
\textbf{The upper part} covers widely used multimodal document understanding datasets; \textbf{the lower part} compiles recent multimodal RAG benchmarks introduced by methods surveyed in this paper to address prior limitations.}
\label{tab:DB_overreview}
\vspace{-6mm}
\end{table}

%% file: section/Application.tex
\section{Application}
Multimodal RAG increasingly serves document understanding across finance, scientific research, and survey analysis. In finance, MultiFinRAG~\citep{gondhalekar2025multifinrag} improves question answering over reports by jointly modeling narrative text, tables, and figures, while FinRAGBench-V~\citep{zhao2025finragbench} provides a benchmark that emphasizes visual citation for transparent evidence traceability in financial documents. In the scientific domain, HiPerRAG~\citep{gokdemir2025hiperrag} enables cross-modal retrieval and reasoning at the scale of millions of research papers, and CollEX~\citep{schneider2025collex} supports interactive exploration of multimodal scientific corpora. In the social sciences, a Eurobarometer-based framework embeds RAG with multimodal LLMs~\citep{papageorgiou2025multimodal} to unify text and infographics, improving the interpretability of survey data. Taken together, these applications demonstrate how multimodal RAG strengthens the capacity to understand and leverage complex documents across fields.

\section{Challenge, Critical Analysis and Industry Deployment}
Due to space constraints, extended discussions are deferred to the appendix. Appendix~\ref{sec:challenge} outlines key open challenges and future directions in multimodal RAG, focusing on efficiency, training paradigms, granularity, and security. Appendix~\ref{sec:critical_analysis} presents a concise critical analysis of fundamental limitations and representative failure cases beyond aggregate benchmarks. Appendix~\ref{sec:industrial-deployment} addresses industrial deployment considerations, highlighting practical constraints, efficiency trade-offs, and representative open-source systems.

%% file: section/Conclusion.tex
\section{Conclusion}
This survey provides a systematic overview of multimodal RAG for document understanding. We analyze methodological advances across retrieval modalities, domain settings, retrieval granularity, and the incorporation of graph-based and agent-oriented architectures, highlighting how these developments enhance understanding over visually rich documents. We further consolidate key datasets, benchmarks, and applications in finance, scientific literature, and social analysis, illustrating the broad impact of multimodal RAG. Despite these advances, challenges remain in efficiency, fine-grained multimodal representation, and robustness in real-world deployment. Addressing these issues will be crucial for future advancement, and we hope this work provides a foundation for advancing multimodal RAG toward reliable and generalizable document AI.




%% file: section/Limitation.tex
\section*{Limitations}
Although this survey aims to provide a comprehensive synthesis of multimodal RAG for document understanding, several limitations remain.
First, while we highlight practical applications, our analysis of real-world deployment challenges such as 
user-centered evaluation, system integration, and deployment scalability remains preliminary. Broader socio-technical aspects of multimodal RAG systems deserve further exploration in future work.
Second, although we summarize major datasets and benchmarks, a more systematic investigation into data quality, annotation consistency, inter-domain transferability, and evaluation alignment across modalities would provide deeper insights into their generalizability and real-world relevance.
Furthermore, as multimodal RAG for document understanding is an emerging and rapidly evolving field, newly released datasets, models, and evaluation protocols continue to reshape the landscape. To address this dynamic nature, this survey will be periodically updated and complemented by an open repository to track ongoing progress and facilitate community collaboration.

\section*{Ethics Statement}
Our work is a survey of existing literature and does not introduce new models, algorithms, or datasets. Therefore, the survey itself does not create new risks. However, we acknowledge that the technologies we review, \textit{i.e.,} multimodal RAG for document understanding, have some potential risks: 1) bias and discrimination inherited from the training data, and 2) the generation of misinformation due to model hallucination. We highlight that addressing these ethical challenges is a critical direction for future research.

\textbf{The Use of AI assistants.} AI assistants (ChatGPT) are used to correct potential grammatical inaccuracies in the manuscript. AI assistants do not participate in research ideation.

%% file: appendix/datasets.tex
\section{Dataset and Benchmark}
\label{sec:appendix_db}
\input{table/benchmark_all}

In the main body, we provide a systematic introduction to the datasets and benchmarks that are widely used for multimodal RAG in document understanding. For each dataset or benchmark, we include a more detailed description, as summarized in Table~\ref{tab:appendix_db}, which lists the data sources and key characteristics. For instance, DocVQA~\citep{mathew2021docvqa} is derived from the UCSF Industry Collections, InfoVQA~\citep{mathew2022infographicvqa} originates from diverse infographics, and TAT-DQA~\citep{zhu2021tat} is constructed from financial reports containing semi-structured tables and text.

In addition, we compile the evaluation results of various multimodal RAG methods on widely used benchmarks, including DocVQA~\citep{mathew2021docvqa}, InfoVQA~\citep{mathew2022infographicvqa}, SlideVQA~\citep{tanaka2023slidevqa}, and MMLongBench-Doc~\citep{ma2024mmlongbench}, as presented in Table~\ref{tab:rag_eval}. These results provide a clear comparison of the strengths and weaknesses of different approaches. The evaluation of multimodal RAG performance typically falls into two categories: retrieval and generation, which are presented in the upper and lower parts of Table~\ref{tab:rag_eval}, respectively. Retrieval evaluation focuses on the accuracy of the retrieved pages, whereas generation evaluation measures the correctness of model outputs when the retrieved pages are combined with the user query as input. Since different methods adopt slightly different metrics, we annotate these variations in the table, while aligning comparable metrics to facilitate direct comparison. Detailed explanations of these metrics are provided in  Appendix~\ref{sec:appendix_eval}.

%% file: table/benchmark_all.tex
\begin{table*}[t]
\centering
\setlength{\tabcolsep}{16pt}
\resizebox{\linewidth}{!}{
\begin{tabular}{l l c c c c}
\toprule
\textbf{Method} & \textbf{Metric} & \textbf{DocVQA} & \textbf{SlideVQA} & \textbf{InfoVQA} & \textbf{MMLongBench-Doc} \\
\midrule
\rowcolor{blue!10} \multicolumn{6}{l}{\textbf{Retrieval Evaluation}} \\
SV-RAG~\citep{chen2024sv}       & Top-5   & 87.0  & 98.8  & --    & 84.8 \\
DSE~\citep{ma2024unifying}            & R@10    & --    & 84.6  & --    & -- \\
VisRAG~\citep{yu2024visrag}         & R@10    & 91.20 & 97.39 & 97.08 & -- \\
CMRAG~\citep{chen2025cmrag}          & R@10    & --    & --    & --    & 64.12 \\
RegionRAG~\citep{li2025regionrag} & R@10 & 99.4 & 98.4 & 99.5 & -- \\
VisRAG~\citep{yu2024visrag}         & MRR@10  & 75.37 & 91.85 & 86.37 & -- \\
CMRAG~\citep{chen2025cmrag}          & MRR@10  & --    & --    & --    & 47.64 \\
LILaC~\citep{yun2025lilac} & MRR@10 &  78.75 & 84.43 &  86.83 & -- \\
ColPali~\citep{faysse2024colpali}        & nDCG@5  & 54.4  & --    & 81.8  & -- \\
ColQwen2~\citep{faysse2024colpali}       & nDCG@5  & 61.5  & --    & 89.7  & -- \\
ColQwen2.5~\citep{faysse2024colpali}     & nDCG@5  & 63.6  & --    & 92.5  & -- \\
VDocRAG~\citep{tanaka2025vdocrag}        & nDCG@5  & --    & 77.3  & 72.9  & -- \\
Light-ColPali~\citep{ma2025towards}  & nDCG@5  & 53.4  & 91.7  & 82.8  & 73.3 \\
Light-ColQwen2~\citep{ma2025towards} & nDCG@5  & 56.6  & 92.9  & 89.5  & 77.0 \\
RegionRAG~\citep{li2025regionrag} & nDCG@5 & 93.1 & 90.3 & 94.8 & -- \\
HKRAG~\citep{tong2025hkrag} & nDCG@5 & -- & 74.3 & 71.9 & -- \\
DSE~\citep{ma2024unifying}            & nDCG@10 & --    & 75.3  & --    & -- \\
CMRAG~\citep{chen2025cmrag}          & nDCG@10 & --    & --    & --    & 52.10 \\
\midrule
\rowcolor{green!10} \multicolumn{6}{l}{\textbf{Generation Evaluation}} \\
VisRAG~\citep{yu2024visrag}       & EM    & 67.17 & 60.97 & 66.43 & -- \\
FRAG~\citep{huang2025frag}         & EM         & --    & 72.7  & --    & -- \\
LILaC~\citep{yun2025lilac} & EM & 65.48 & 55.57 & 60.91 & -- \\ 
SV-RAG~\citep{chen2024sv}       & PNLS & 76.0 & 77.0 & -- & 49.0 \\
CRAEM~\citep{zhang2024cream}       & ANLS       & 79.4  & --    & 53.6  & -- \\
M3DocRAG~\citep{cho2024m3docrag}     & ANLS       & 84.4  & --    & --    & -- \\

VisDoMRAG~\citep{suri2025visdom}    & ANLS       & --    & 67.2  & --    & -- \\
VDocRAG~\citep{tanaka2025vdocrag}      & ANLS       & --    & 56.4  & 64.6  & -- \\
FRAG~\citep{huang2025frag}         & ANLS       & 87.4  & --    & --    & -- \\
ReDocRAG~\citep{lopez2025enhancing}    & ANLS       & 73.7  & --    & 63.6  & -- \\
M3DocRAG~\citep{cho2024m3docrag}     & G-Acc        & --    & --    & --    & 21.0 \\
FRAG~\citep{huang2025frag}         & G-Acc        & 80.5  & --    & --    & 37.9 \\
VRAG-RL~\citep{wang2025vrag}      & G-Acc        & --    & 69.3  & --    & 24.9 \\
SimpleDoc~\citep{jain2025simpledoc}    & G-Acc        & --    & --    & --    & 60.58 \\
MMRAG-DocQA~\citep{gong2025mmrag} & G-Acc        & --    & --    & --    & 52.3 \\
CMRAG~\citep{chen2025cmrag}        & G-Acc        & --    & --    & --    & 43.25 \\
MoLoRAG~\citep{wu2025molorag}      & G-Acc        & --    & --    & --    & 41.01 \\
RECON~\citep{wangrecon} & G-Acc & -- & 66.12 & -- & -- \\
LAD-RAG~\citep{sourati2025lad} & G-Acc & 82.9 & -- & -- & 45.0 \\
DREAM~\citep{zhang2025dream} & G-Acc & -- & -- & -- & 27.3 \\
MARA~\citep{wu2025mara} & G-Acc & 84.64 & 73.40 & 68.02 & -- \\
SLEUTH~\citep{liu2025resolving} & G-Acc & -- & -- & -- & 52.77 \\
\bottomrule
\end{tabular}}
\caption{Evaluation results of RAG methods. The upper block shows \textbf{retrieval evaluation} and the lower block shows \textbf{generation evaluation}. Different background shades are used to separate the two parts.}
\label{tab:rag_eval}
\end{table*}

%% file: appendix/eval.tex
\section{Evaluation Metrics}
\label{sec:appendix_eval}
The evaluation of multimodal RAG methods typically involves two aspects: retrieval evaluation and generation evaluation.
Retrieval primarily measures the system’s ability to accurately retrieve relevant multimodal information from a large corpus.
Generation, on the other hand, evaluates the quality of the model’s produced outputs conditioned on the retrieved context.
We list the most commonly used metrics along with some newly designed ones that address the limitations in the following.

\subsection{Retrieval Evaluation}
In the context of multimodal RAG, a variety of metrics are commonly employed to evaluate the performance of the retriever module. Popular measures include Accuracy, Recall, Precision, F1-Score~\citep{christen2023review}, Mean Reciprocal Rank (MRR)~\citep{omar-etal-2024-multi, nguyen2024}, and Normalized Discounted Cumulative Gain (nDCG)~\citep{jarvelin2002cumulated}.

A widely used measure is Top-K Accuracy, which reflects the hit rate of retrieval. 
\begin{equation}
\begin{aligned}
\text{Top-}K\;\text{Accuracy} 
&= \frac{1}{|Q|} \sum_{q \in Q} \\
&\quad \mathbf{1}\Bigl( \mathrm{Rel}(q) 
    \cap \mathrm{Ret}_K(q) \neq \varnothing \Bigr),
\end{aligned}
\end{equation}
where, for a given query q, $\mathrm{Ret}_K(q)$ denotes the set of the top-K results returned by the retrieval system, $\mathrm{Rel}(q)$ denotes the set of all ground-truth relevant documents or modality segments, and Q denotes the collection of queries. The same symbols appearing in the following formulas carry the same meanings.

Recall@K is usually used to quantify retrieval coverage, measuring how many of the ground-truth relevant items are captured within the top K results:
\begin{equation}
\mathrm{Recall}@K = \frac{1}{|Q|} \sum_{q \in Q} \frac{\lvert \mathrm{Rel}(q) \cap \mathrm{Ret}_K(q) \rvert}{\lvert \mathrm{Rel}(q) \rvert}.
\end{equation}

Precision@K instead measures accuracy, i.e., the proportion of retrieved items among the top K that are relevant:
\begin{equation}
\mathrm{Precision}@K = \frac{1}{|Q|} \sum_{q \in Q} \frac{\lvert \mathrm{Rel}(q) \cap \mathrm{Ret}_K(q) \rvert}{K}.
\end{equation}

The F1-Score is often adopted as the harmonic mean of Precision@K and Recall@K, widely used to assess the correctness of retrieved entities or factual fragments in both the retrieval module and the generation process~\citep{li2024benchmarking}:
\begin{equation}
\mathrm{F1}@K = \frac{1}{|Q|} \sum_{q \in Q} 2 \cdot \frac{\mathrm{Pr}_K(q) \cdot \mathrm{Re}_K(q)}{\mathrm{Pr}_K(q) + \mathrm{Re}_K(q)},
\end{equation}
where, $\mathrm{Pr}_K(q)$ represents Precision@K, and $\mathrm{Re}_K(q)$ represents Recall@K.
%



However, the metrics above are insensitive to the ranking order within the top $K$.
In practice, placing highly relevant or informative items at earlier positions is crucial for effective RAG.
\citet{omar-etal-2024-multi, nguyen2024} utilize MRR@K to emphasize the position of the first relevant item:
\begin{equation}
\mathrm{MRR}@K = 
\frac{1}{|Q|}
\sum_{q \in Q}
\frac{\mathbf{1}\bigl(\mathrm{rank}_K(q) \le K\bigr)}{\mathrm{rank}_K(q)},
\end{equation}
where $\mathrm{rank}_K(q)$ denotes the position of the first relevant document within the top-$K$ retrieved results for query $q$;
if no relevant item appears within the top $K$, the reciprocal rank is set to $0$.

Similarly, \citet{zhao2025finragbench,faysse2024colpali} employ nDCG@K to penalize relevant items that appear lower in the ranking, thereby rewarding systems that surface high-quality evidence earlier:
\begin{equation}
\mathrm{nDCG}@K = 
\frac{1}{|Q|} \sum_{q \in Q}
\frac{\mathrm{DCG}@K(q)}{\mathrm{IDCG}@K(q)},
\end{equation}
where
\begin{equation}
\mathrm{DCG}@K(q) = 
\sum_{i=1}^{K} 
\frac{2^{\mathrm{rel}_{q,i}} - 1}{\log_2(i + 1)}.
\end{equation}
Here, $\mathrm{rel}_{q,i}$ represents the graded relevance of the $i$-th retrieved item for query $q$. The denominator $\mathrm{IDCG}@K(q)$, called the \textit{ideal DCG}, represents the maximum possible DCG that could be achieved for query $q$ if all relevant items were perfectly ranked at the top of the list.

\subsection{Generation Evaluation}
%
In the context of Multimodal RAG, the primary objective of generation quality evaluation is to assess the quality and consistency between model-generated text and reference answers. This involves not only measuring the correctness of the responses but also considering aspects such as fluency, information coverage, and logical coherence. To achieve a comprehensive evaluation, this study examines a wide range of metrics. The earliest are soft matching metrics (\emph{e.g.}, BLEU~\citep{papineni2002bleu}, ROUGE~\citep{lin2004rouge}, METEOR~\citep{banerjee2005meteor}), which rely on n-gram overlap for a soft lexical evaluation that allows partial and flexible matching. They mainly assess fluency and information coverage of generated text.
With the rise of question answering and reading comprehension tasks, strict matching metrics (\emph{e.g.}, Exact Match~\citep{rajpurkar-etal-2016-squad}, ANLS~\citep{biten2019scene}, PNLS~\citep{chen2024mmr}) are introduced, focusing on exact or near-exact correspondence with reference answers to measure form-level correctness. 
More recently, driven by the advances in pretrained language models, semantic matching metrics (\emph{e.g.}, BERTScore~\citep{devlin2019bert}, RoBERTa~\citep{liu2019roberta}, G-Acc~\citep{ma2024mmlongbench}) have become prominent, enabling the assessment of deeper semantic consistency through contextual embeddings. By combining these three categories of metrics, generation quality can be evaluated holistically across surface, exact matching, and semantic alignment.

\paragraph{Soft Matching Metrics.}
The earliest approaches to generation quality evaluation adopt soft matching metrics, which rely on n-gram overlap to provide a soft lexical evaluation that tolerates partial and flexible matching between generated and reference texts. Among them, BLEU~\citep{papineni2002bleu} is one of the most representative metrics. BLEU evaluates the similarity between generated text and reference text based on n-gram overlap with a brevity penalty (BP). The BLEU score is defined as:
\begin{equation}
\text{BLEU} = \text{BP} \cdot \exp\left(\sum_{n=1}^N w_n \log p_n \right),
\end{equation}
where $p_n$ is the precision for n-grams and $w_n$ is the weight assigned to each n-gram order.
The brevity penalty (BP) is given by:
\begin{equation}
\text{BP} = \exp\!\left( \min\!\left(0,\, 1 - \frac{r}{c}\right) \right),
\end{equation}
where c is the candidate (generated) length and r is the reference length.

Compared to BLEU~\citep{papineni2002bleu}, ROUGE~\citep{lin2004rouge} evaluates the overlap between generated and reference texts at the n-gram level, and is widely used in summarization tasks. The ROUGE-N score is defined as:
\begin{equation}
\text{ROUGE-}N = \frac{\sum_{\text{ref}} \sum_{n \in \text{ref}} \min\bigl(S_n, R_n\bigr)}{\sum_{\text{ref}} \sum_{n \in \text{ref}} R_n},
\end{equation}
where $S_n$  and $R_n$ denote the counts of a given n-gram in the system output and reference, respectively.
ROUGE-L leverages the Longest Common Subsequence (LCS) between the system output and the reference to capture sentence-level structural similarity. Its recall-oriented form is given by:
\begin{equation}
\text{ROUGE-L} = \frac{\text{LCS}(S, R)}{|R|},
\end{equation}
where \text{LCS}(S, R) denotes the length of the longest common subsequence between the system output S and the reference R, and |R| is the length of the reference.

Compared to BLEU and ROUGE, METEOR~\citep{banerjee2005meteor} emphasizes semantic matching beyond exact n-gram overlap. It incorporates stemming, synonym matching, and a penalty for word order differences to better capture the similarity between system outputs and references. The METEOR score is defined as:
\begin{equation}
\text{METEOR} = F_{\alpha} \cdot (1 - P),
\end{equation}
where $F_{\alpha}$ is a weighted harmonic mean of precision ($P_{pre}$) and recall ($P_{rec}$), given by:
\begin{equation}
F_{\alpha} = \frac{P_{rec} \cdot P_{pre}}{\alpha \cdot P_{pre} + (1-\alpha) \cdot P_{rec}},
\end{equation}
and $P$ is a fragmentation penalty based on word order:
\begin{equation}
P = \gamma \left( \frac{ch}{m} \right)^{\beta},
\end{equation}
where $ch$ denotes the number of chunks (i.e., contiguous matched word sequences), $m$ is the total number of matched words, and $\alpha, \beta, \gamma$ are tunable parameters.

\paragraph{Strict Matching Metrics.}
In contrast to soft matching metrics, strict matching metrics emphasize exact or near-exact correspondence between generated and reference answers.
They assess the consistency and form-level correctness of model outputs, directly reflecting the factual accuracy of the generated responses.

The most representative metric in this category is Exact Match (EM)~\citep{rajpurkar-etal-2016-squad}, which computes the percentage of predictions that exactly match one of the reference answers:  
\begin{equation}
\mathrm{EM} = \frac{1}{N} \sum_{i=1}^{N} \mathbf{1}(o_i \in A_i),
\end{equation}
where $o_i$ denotes the predicted answer, $A_i$ is the set of groundtruth answers, and $\mathbf{1}(\cdot)$ is the indicator function.  

With the advancement of generative models and their increasing generalization capabilities, more recent metrics have been introduced. Average Normalized Levenshtein Similarity(ANLS)~\citep{biten2019scene} is designed to provide a soft evaluation of string-based answers. ANLS is defined as below:

\begin{equation}
\mathrm{NLS}(a_{ij}, o_i) = 1 - \frac{\mathrm{LD}(a_{ij}, o_i)}{\max(|a_{ij}|, |o_i|)},
\end{equation}

\noindent where $o_i$ is a given prediction, $a_{ij}$ is a groundtruth answer, $\mathrm{LD}(a_{ij}, o_i)$ denotes the standard Levenshtein edit distance~\citep{lcvenshtcin1966binary}, and $|\cdot|$ is the string length.
The threshold $\tau$ controls the minimum similarity required for a prediction to be considered correct.
\begin{equation}
s(a_{ij}, o_i) =
\begin{cases}
\mathrm{NLS}(a_{ij}, o_i), & \text{if } \mathrm{NLS}(a_{ij}, o_i) \geq \tau, \\
0, & \text{otherwise},
\end{cases}
\end{equation}
\begin{equation}
\mathrm{ANLS} = \frac{1}{N} \sum_{i=1}^N \max_j s(a_{ij}, o_i).
\end{equation}

Moreover, AccANLS~\citep{zhang2024exploring} integrates accuracy with ANLS similarity, aiming at addressing the issue of penalizing redundant outputs. Partial Normalized Levenshtein Similarity (PNLS)~\citep{chen2024mmr} generalizes ANLS by relaxing the alignment requirement: instead of computing edit distance over the entire strings, it identifies the best-matching substring of the prediction relative to the reference. This design avoids penalizing extra prefixes or suffixes while still accounting for mismatches, insertions, and deletions within the aligned region, making it more suitable for evaluating verbose LLM outputs. Formally, PNLS still follows the NLS formulation but replaces the standard edit distance with a \emph{partial edit distance} $\mathrm{LD}^*(a_{ij}, o_i)$ obtained via approximate string matching~\citep{sellers1980theory}. The final score is computed as:  
\begin{equation}
\mathrm{PNLS}(a_{ij}, o_i) = 1 - \frac{\mathrm{LD}^*(a_{ij}, o_i)}{\max(|a_{ij}|, |o_i'|)},
\end{equation}
where $o_i'$ denotes the optimally aligned substring of the prediction $o_i$.  

\paragraph{Semantic Matching Metrics.}
Beyond soft and strict matching metrics, semantic matching metrics have emerged to evaluate deeper semantic consistency between generated and reference texts. Metrics such as BERTScore, which leverages contextual embeddings from pretrained language models like BERT~\citep{devlin2019bert} and RoBERTa~\citep{liu2019roberta}, move beyond simple lexical overlap by capturing semantic similarity between generated and reference texts. This enables a more reliable evaluation of whether the meaning of a response is preserved, even when different phrasings are used. However, while BERTScore provides strong advantages in measuring semantic consistency, it is less suited for scenarios involving long-form, explanatory, or unanswerable responses. To address this gap, Generated Accuracy (G-Acc)~\citep{ma2024mmlongbench} has been proposed, which extends evaluation to free-form answers that emphasize reasoning, elaboration, and contextual completeness, thereby offering a more comprehensive assessment of generation quality.

%% file: appendix/loss.tex
\section{Training Loss}
\label{sec:appendix_loss}
In multimodal RAG, the most common training objective is a ColBERT-style~\citep{khattab2020colbert,faysse2024colpali} contrastive loss. The key idea is to represent both queries and documents with multiple contextualized token embeddings and compute their similarity through a \textit{late interaction mechanism}. Formally, given a query $q$ and a document $d$, we represent them as
$\mathbf{H}_q \in \mathbb{R}^{L_q \times D}$ and $\mathbf{H}_d \in \mathbb{R}^{L_d \times D}$,
where $L_q$ and $L_d$ denote the number of tokens in the query and document, and $D$ is the embedding dimension.
The late interaction similarity is defined as:
\begin{equation}
\text{Sim}(q, d) = \sum_{t=1}^{L_q} \max_{1 \le m \le L_d} 
\left\langle \mathbf{h}_{q,t}, \mathbf{h}_{d,m} \right\rangle,
\end{equation}
where $\langle \cdot , \cdot \rangle$ denotes the dot product. This operator allows each query token to attend to its most relevant document token, enabling fine-grained matching.

During training, a contrastive objective~\citep{khosla2020supervised,wang2021understanding} is optimized over a batch of query–document pairs $\{ (x_i, y_i) \}_{i=1}^{B}$.
For each query $x_i$, the paired document $y_i$ is the positive example, while the remaining documents in the batch act as negatives.
Let $p_i = \text{Sim}(x_i, y_i)$ and $n_i = \max_{j \neq i} \text{Sim}(x_i, y_j)$ denote the positive and hardest negative similarities, respectively.
The loss is defined as:
\begin{equation}
\begin{split}
\mathcal{L} &= -\frac{1}{B} \sum_{i=1}^B 
\log \left( \frac{\exp(p_i)}{\exp(p_i) + \exp(n_i)} \right) \\
&= \frac{1}{B} \sum_{i=1}^B 
\log \left( 1 + \exp(n_i - p_i) \right),
\end{split}
\end{equation}
which encourages higher similarity for the positive pair than for any in-batch negative.

This ColBERT-style loss, combining late interaction with contrastive learning, is widely adopted in multimodal RAG systems as it provides effective supervision for aligning queries and documents across both text and vision modalities.

%% file: section/Challenge.tex
\section{Challenge and Future Direction}
\label{sec:challenge}

Although multimodal RAG has made continuous progress in the field of document understanding, there are still several key challenges. Future research mainly focuses on the following aspects: efficiency, document-specific model architectures and training paradigms, granular and scalable evaluation protocols, and security and robustness for high-risk application scenarios. 

\paragraph{Model Architectures, Training Paradigms, and Efficiency.}
The current VLMs~\citep{bai2023qwen,chen2024internvl,beyer2024paligemma} are mainly designed for general image-text benchmarks and lack specialized architectures for the unique visual structures in documents (such as diagrams, icons, tables, and complex formulas). This often leads to the inability to fully preserve fine-grained layout information and symbolic cues, prompting researchers to explore domain-specific vision encoders to better capture the structural and semantic features crucial for document understanding. In terms of training paradigms, many retrieval systems adopt the late interaction mechanism of ColBERT-style~\citep{khattab2020colbert,faysse2024colpali,masry2025colmate}. One core limitation of this design lies in its scalar scoring method based on MaxSim, which only focuses on the most similar token pairs and ignores the broader semantic alignment relationships between tokens. Therefore, in semantic-rich document scenarios, it is difficult to capture distributed and subtle correlation signals. Future research can alleviate this problem by exploring more comprehensive token interaction goals beyond simple maximum aggregation.
Efficiency is one of the core challenges of multimodal retrieval systems, especially in scenarios where thousands or even millions of documents need to be processed. Encoding based on VLMs generates a large number of visual tokens for each document (see Table~\ref{tab:bench_tokens}), significantly increasing storage and retrieval computational costs. Techniques such as token compression, visual token merging, and dynamic pruning provide feasible paths to reduce this burden~\citep{ma2025towards,kim2025hybrid,bach2025hierarchical}. However, effectively reducing computational costs without significantly compromising retrieval performance remains an important direction for future research. 

\paragraph{Granular Understanding and Evaluation Protocols.}
More granular document representation is necessary~\citep{wang2025vrag,xu2025multi,yu2025beyond,gong2025mmrag,choi2025zero,zhang2023ideal,wang2022cris}. This is because many existing models still operate at the page-level modeling, ignoring key elements such as tables, figures, footnotes, and layout-specific semantics. However, the progress in this direction is severely limited by the current benchmarks and scoring functions~\citep{faysse2024colpali,mathew2022infographicvqa,ma2024mmlongbench}. Existing datasets usually rely on single-hop retrieval on small-scale corpora and cannot effectively test the scalability or retrieval accuracy of the system. There is an urgent need to build an open-domain benchmark containing thousands of mixed-modal documents to evaluate the needle-in-a-haystack retrieval capability. Such a benchmark needs to focus on testing the model's ability to locate specific visual elements, rather than simply retrieving relevant pages~\citep{yu2025bbox}. At the same time, standard metrics such as Recall@K treat pages as atomic units, which are not precise enough in multimodal scenarios because a single page often contains multiple independent information sources. We propose to introduce hierarchical metrics and visual grounding scores~\citep{liu2024grounding,deng2021transvg,xiao2024towards}, which focus on retrieving specific visual evidence (such as a particular table or chart), rather than the entire page content, thereby improving the interpretability of the evaluation and supporting more complex downstream inference tasks. 

\paragraph{Security, Robustness, and Trust.}
With the widespread deployment of multimodal RAG systems in high-risk fields such as finance, healthcare, and law, security and robustness have become critical issues that cannot be ignored~\citep{shereen2025one,cho2024typos,nazary2025poison,jiang2024rag,xian2024understanding,gao2024boosting,jia2025adversarial,jia2025semantic,xiang2026safety}. Besides hallucination and data leakage, the multimodal scenario also introduces cross-modal attack surfaces. Attackers can manipulate retrieval results through adversarial images, layouts, or visual cues, or guide the generation model to produce incorrect legal, medical, or financial conclusions, even bypassing text-based security filtering mechanisms~\citep{abootorabi2025ask,liu2025poisoned}. At the same time, most existing systems lack mechanisms for cross-modal verification of retrieval and generation of evidence sources (provenance), making targeted knowledge poisoning difficult to detect. Therefore, reliable deployment requires the introduction of privacy-preserving retrieval, verifiable generation, and risk-aware trust calibration, and the design of evaluation protocols that go beyond accuracy metrics to systematically assess the robustness of the model in adversarial and poisoning attack scenarios~\citep{nazary2025poison}.

%% file: appendix/hybrid.tex
\section{Critical Analysis}
\label{sec:critical_analysis}

While recent methods have achieved notable gains in Multimodal RAG benchmarks, a closer examination reveals several unresolved contradictions that are often obscured by aggregate performance improvements. In this section, we critically analyze prevailing paradigms, focusing on the tension between visual and textual representations, the robustness of evaluation protocols, and the trade-offs between system complexity and practical utility.

\paragraph{The "OCR-Free" vs. "OCR-Based" Paradox.}
A growing body of work (\textit{e.g.}, ColPali~\citep{faysse2024colpali}, VisRAG~\citep{yu2024visrag}) promotes OCR-free approaches that encode document pages directly using vision–language models, thereby avoiding error propagation introduced by OCR systems. While such methods are effective at capturing layout structure and visual elements such as tables and charts, they remain vulnerable to visual hallucination when handling dense, fine-grained text or precise numerical information, as commonly found in financial and technical documents~\citep{maleki2024ai,liu2024survey,wang2024mitigating}. In contrast, OCR-based pipelines sacrifice certain layout semantics but typically offer higher fidelity for text-centric retrieval tasks, particularly those requiring exact string matching or keyword search. Despite this, recent literature often frames OCR-free methods as a universal progression, overlooking their persistent weaknesses in text-intensive scenarios. This unresolved dichotomy highlights the absence of a unified representation that can simultaneously preserve visual structure and ensure symbolic precision, underscoring a fundamental limitation in current Multimodal RAG systems.

\paragraph{Validity and Saturation of Current Benchmarks.} The rapid saturation of performance on standard benchmarks (such as DocVQA~\citep{mathew2021docvqa} and InfoVQA~\citep{mathew2022infographicvqa}) raises concerns about their validity as proxies for real-world document understanding. First, data contamination is a significant, often unaddressed risk. Given that many LLMs are pre-trained on vast web corpora, there is a non-negligible possibility that public benchmark data has leaked into the training sets, rendering high scores indicative of memorization rather than reasoning~\citep{xu2024benchmarking,hu2025vlsbench,zhou2025lessleak,xu2024benchmark,deng2024investigating}. Second, there is a misalignment between benchmark tasks and practical RAG scenarios. Most existing datasets focus on single-page or short-document VQA. However, the core challenge of Multimodal RAG lies in retrieving the correct needle from a haystack of thousands of pages~\citep{faysse2024colpali,tanaka2025vdocrag}. High performance on current generation-focused benchmarks does not necessarily translate to robustness in large-scale, open-domain retrieval settings.

\paragraph{The Complexity–Performance Trade-off.} Recent work increasingly adopts complex mechanisms such as graph-based indexing~\citep{liu2025hm,yuan2025mkg,wangrecon,sourati2025lad}, agentic workflows~\citep{liu2025resolving,chen2025hear,zhang2025patho}, and multi-round self-reflection. However, these designs often lead to only marginal performance gains (e.g., a 1–2\% increase in accuracy) while significantly increasing computational overhead and inference latency. Despite this imbalance, few studies provide a clear cost–benefit analysis. For example, agent-based methods that require multiple LLM calls per query may be novel from a research perspective, but they are often impractical for real-time industrial deployment compared to simpler, well-tuned dense retrieval baselines. Overall, the literature tends to favor architectural complexity while paying insufficient attention to the resulting costs in latency and token usage.

\section{Industry Deployment}
\label{sec:industrial-deployment}
The main body of this survey focuses on model architectures, retrieval paradigms, datasets, and benchmarks for multimodal RAG in document understanding. Beyond methodological advances, industrial deployment plays a critical role in determining the practical impact of these systems. In real-world settings, multimodal RAG is primarily applied to large-scale industrial documents, where efficiency, reliability, and system integration are central concerns. Accordingly, this section discusses multimodal RAG from an industry perspective, with a focus on industrial document characteristics, efficiency considerations in retrieval systems, and representative open-source tools that facilitate practical deployment.

\FloatBarrier
\begin{table*}[t]
\centering
\setlength{\tabcolsep}{4pt}
\resizebox{\linewidth}{!}{
\begin{tabular}{lcp{12cm}}
\hline
Project & Stars (Dec. 2025) & Key features \\
\hline
RAGFlow~\citep{ragflow} & $\sim$70.3k &
Enterprise-grade RAG engine with agents, document processing (DeepDoc), graph-based retrieval, and rich UI for production deployments. \\
RAG-Anything~\citep{RAG-Anything} & $\sim$11.3k &
``All-in-one'' RAG framework with modular pipelines, multi-backend support, and examples for text and document QA. \\
LightRAG~\citep{LightRAG} & $\sim$26.6k &
Simple and fast RAG with graph-enhanced retrieval, scalable to large corpora and offering Docker/K8s deployment recipes. \\
AutoRAG~\citep{AutoRAG} & $\sim$4.5k &
AutoML-style framework for RAG evaluation and optimization, automatically exploring retrievers, chunkers, and generators. \\
RAGLite~\citep{raglite} & $\sim$1.1k &
Lightweight Python toolkit that implements RAG directly over DuckDB or PostgreSQL, emphasizing simplicity and SQL-native integration. \\
LlamaIndex~\citep{LlamaIndex} & $\sim$46k &
General framework for building RAG and agentic systems over arbitrary data sources, with extensive connectors and ecosystem. \\
\hline
\end{tabular}
}
\caption{Representative open-source RAG frameworks frequently used in industrial-style deployments. GitHub star counts are approximate and reported as of December~2025.}
\label{tab:industrial-rag-frameworks}
\end{table*}

\paragraph{Domain-specific multimodal RAG in industry.}
RAG has been widely applied to industrial knowledge bases~\citep{riedler2024beyond,liu2024optimizing,bourdin2025agile,brehme2025retrieval,chen2025application}. In industrial knowledge management, RAG systems support tasks such as troubleshooting, maintenance, and querying internal regulations, with an emphasis on domain-specific indexing and user-oriented workflows rather than model innovation. In manufacturing, RAG is also integrated into cognitive digital twin systems~\citep{shi2025enhancing}, where it operates over structured asset representations such as Asset Administration Shells to support tasks including system integration and model alignment. More recently, Golden-Retriever~\citep{an2024golden} explores agentic RAG for industrial knowledge bases by combining high-quality retrieval, re-ranking, and tool-using agents to enable multi-step reasoning and coordinated data access.

Multimodal RAG is particularly suitable for industrial document analysis scenarios. In such scenarios, long documents typically contain text, tables, and charts, and have high requirements for processing efficiency and reliability in actual deployments. Financial documents can be regarded as a typical form of industrial documents, with similar characteristics in terms of structural complexity and engineering constraints. Driven by benchmarks such as TAT-DQA~\citep{zhu2021tat} and FinRAGBench-V~\citep{zhao2025finragbench}, recent research has begun to focus on conducting question answering on long, visually rich documents. Systems like MultiFinRAG~\citep{gondhalekar2025multifinrag} enhance retrieval effectiveness by jointly indexing multiple modalities. IndustryRAG~\citep{lim2025distilling} further emphasizes efficiency by distilling domain knowledge and structural knowledge into a compact retriever, making multimodal RAG more practical for industrial deployments. 

\paragraph{Efficiency of visual embeddings for large-scale deployment.}
Industrial corpora typically consist of thousands of multi-page PDFs, scanned manuals, CAD-like drawings, and complex financial charts. Storing dense visual embeddings for each page or every visual element would quickly become unmanageable in terms of memory usage and retrieval latency. Efficiency-oriented approaches~\citep{ma2025towards,yan2025docpruner,bach2025hierarchical,kim2025hybrid} (such as Light-ColPali~\citep{ma2025towards}) alleviate this bottleneck by compressing page-level visual representations. Light-ColPali reduces the number of patch-level embeddings through token merging while retaining the late-interaction scoring mechanism, achieving near-optimal retrieval quality with only a small portion of the original visual tokens. From a deployment perspective, these methods significantly reduce GPU memory usage and vector-store size, making it possible to index a complete industrial document collection rather than being limited to a small, carefully selected subset. When combined with a closed-domain multimodal RAG strategy and performing only the most relevant page retrieval within the document, visual embedding compression provides a practical solution for expanding industrial systems under strict latency and cost constraints. 

\paragraph{Systems and open-source tooling for rapid deployment.}
The continuously expanding open-source RAG framework ecosystem has significantly lowered the threshold for industrial deployment by addressing practical issues such as system integration, scalability, and maintainability. The overall situation is shown in Table~\ref{tab:industrial-rag-frameworks}. RAGFlow~\citep{ragflow} is designed for production-ready deployment and provides an end-to-end RAG engine with integrated UI, DeepDoc document processing, graph-based retrieval, and agent support, effectively reducing engineering costs in enterprise environments. LlamaIndex~\citep{LlamaIndex} supports rapid integration with heterogeneous data sources through modular ingestion, indexing, and orchestration components and can well adapt to the highly fragmented data infrastructure in industrial scenarios. RAG-Anything~\citep{RAG-Anything} and LightRAG~\citep{LightRAG} place more emphasis on simplicity and scalability. Among them, LightRAG particularly highlights graph-enhanced retrieval and containerized deployment based on Docker and Kubernetes, facilitating the construction of scalable and reproducible industrial systems. AutoRAG~\citep{AutoRAG} addresses a key challenge in industrial deployment and provides automated evaluation and configuration search (AutoML-style optimization) for retriever, chunker, and generator, supporting systematic tuning in cases of limited engineering resources. In contrast, RAGLite~\citep{raglite} adopts a minimalist design, directly built on DuckDB or PostgreSQL (SQL-native integration), and can naturally integrate into the existing enterprise data stack, significantly simplifying long-term maintenance work.

\paragraph{Discussion and open challenges.}
In industrial deployment scenarios, an effective multimodal RAG not only depends on technical design choices but also on the clear definition of role division, workflow, and information model throughout the system's entire lifecycle. Research and practical experience from industrial practice and deployment-oriented studies indicate that there are still several open challenges that need to be addressed at present. Firstly, the quality of retrieval and generation~\citep{bruckhaus2024rag} needs to align with the actual expectations of domain experts rather than relying solely on general benchmarks for evaluation. Secondly, when indexing sensitive text and visual assets, sound data governance~\citep{muller2025data}, access control, and auditability are indispensable. Thirdly, practical monitoring and error analysis tools are needed to accurately attribute system failures to specific modalities or processing stages. Finally, efficiency-oriented technologies such as visual embedding compression and hierarchical retrieval must strike a balance with the demand for faithful and verifiable reasoning capabilities. Solving these challenges is crucial for advancing multimodal RAG from research prototypes to reliable, industry-grade document AI systems.

\section{Graph-based Multimodal RAG}
\label{appendix:graph}
Before moving on to the multimodal scenario, it is necessary to review how the graph structure is introduced into the traditional, text-centric RAG. Recent review works on graph RAG~\citep{peng2024graph,procko2024graph,zhang2025survey} describe a general process: converting documents or knowledge bases into graph structures, selecting subgraphs or local neighborhoods relevant to the query during the retrieval stage, and generating based on graph-structured evidence rather than flat lists of chunks. Compared to vanilla RAG, this paradigm mainly has two advantages: first, it promotes multi-hop reasoning by explicitly modeling the relationships between evidence; second, by anchoring the output of the LLM on coherent evidence paths that connect originally sparse or distant information, it reduces hallucination~\citep{zhang2025survey}. 

\paragraph{Graph-based Textual RAG.}
One of the important research directions in the field of document-level reasoning focuses on the construction of knowledge graph (KG), which involves decomposing documents into entity-centered graphs to achieve cross-page information association~\citep{wangrecon}. Knowledge graph-augmented generation methods such as SubgraphRAG~\citep{li2024simple}, GRetriever~\citep{he2024g}, and ToG-2~\citep{ma2024think} enhance retrieval effectiveness through subgraph selection, ranking-based retrieval, or by combining dense retrieval with graph reasoning. However, these methods usually rely on manually constructed KGs, which have high construction costs and limited coverage. To address this issue, GraphRAG~\citep{edge2024local} uses LLMs to directly construct graphs from the original text and organizes them through hierarchical community detection, enabling document-level reasoning with higher computational costs. Based on this paradigm, subsequent works further explore different design choices and efficiency trade-offs. GNN-RAG~\citep{mavromatis2024gnn} and GFM-RAG~\citep{luo2025gfm} focus on graph-based retrieval and scoring, respectively, supported by graph neural networks or pretrained graph foundation models for cross-document multi-hop reasoning. To reduce indexing and construction costs, KET-RAG~\citep{huang2025ket} proposes a multi-granular indexing scheme that combines lightweight KG skeletons with less costly text-based graphs. More recent variants, such as LightRAG~\citep{guo2024lightrag} and HippoRAG-2~\citep{gutierrez2025rag}, further enhance scalability and reasoning performance by simplifying graph structures and strengthening passage-level integration. 

Despite these advancements, graph-based RAG is currently mainly limited to text-only scenarios and inherits many of the limitations of textual RAG. Therefore, it is difficult to effectively model multimodal signals such as images, tables, or layouts, which are crucial for reasoning in visually rich documents. 

\paragraph{Graph-based Multimodal RAG.}
Graph-based multimodal RAG extends the principles of graph RAG to visually rich documents by explicitly representing multimodal content as a graph structure for modeling. As shown in Figure~\ref{fig:graph_agent}(a), nodes correspond to atomic content units such as pages, text fragments, images, tables, and layout blocks, while edges are used to encode semantic, spatial, and logical relationships. The retrieval process is expressed as selecting a subgraph related to the query to simultaneously capture key content areas and their interrelationships. Reasoning based on this multimodal graph enables LLM to integrate heterogeneous evidence, achieve finer-grained grounding, and provide more interpretable attributions for cross-modal structures. 

The early graph-based multimodal RAG systems have to some extent instantiated the various design roles of graph RAG. HM-RAG~\citep{liu2025hm} adopts a hierarchical multi-agent architecture, treating the graph database as a retrieval modality and using it in parallel with unstructured text and web sources, and aggregating the results through consistency voting. mKG-RAG~\citep{yuan2025mkg} and DB3Team-RAG~\citep{xia2025db3} align the entities and relations in text and images, explicitly constructing multimodal knowledge graphs, thereby supporting knowledge-intensive visual question answering and domain-specific multi-turn queries. As a complement to the aforementioned knowledge-centered methods, MoLoRAG~\citep{wu2025molorag} pays more attention to the document structure and retrieves coherent page sequences by modeling the logical jump relationships between pages. Recent methods have further elevated the graph structure from an auxiliary retrieval component to a core indexing and reasoning framework. RECON~\citep{wangrecon} constructs a global multimodal document graph by linking text and visual relations within pages and introducing entity connections between pages; while LAD-RAG~\citep{sourati2025lad} and LILaC~\citep{yun2025lilac} emphasize layout-aware and component-level graphs, supporting multi-granular and multi-hop multimodal reasoning through subgraph retrieval using dynamic traversal or late interaction.

\paragraph{Discussion and open challenges.}
A key takeaway is that graph structures offer an effective abstraction for organizing and reasoning over multimodal evidence. By explicitly encoding relations among text, images, tables, and layout components, recent methods show clear advantages over flat multimodal retrieval in supporting multi-hop reasoning, fine-grained grounding, and more interpretable evidence aggregation~\citep{edge2024local,wangrecon,sourati2025lad,yun2025lilac}. Nevertheless, constructing reliable multimodal graphs remains nontrivial. Cross-modal alignment and layout relation extraction are often noisy and expensive, and inaccuracies at the graph construction stage can propagate to retrieval and generation, limiting robustness~\citep{yuan2025mkg,xia2025db3}.

Scalability and evaluation pose additional challenges. Large, global multimodal graphs are costly to build and traverse, motivating lightweight indexing schemes and dynamic subgraph retrieval as practical compromises~\citep{huang2025ket,guo2024lightrag}. More generally, existing systems assign very different roles to graphs, ranging from auxiliary retrieval signals to central reasoning scaffolds~\citep{liu2025hm}, suggesting that clearer design principles are needed. Promising directions include adaptive graph construction that adjusts granularity based on query complexity, and hybrid pipelines that combine coarse text retrieval with on-demand multimodal graph reasoning. Finally, progress will require standardized benchmarks and metrics that jointly evaluate graph quality, cross-modal reasoning, and attribution, in order to assess generalization beyond narrow, domain-specific settings.

\section{Agent-based Multimodal RAG}
\label{appendix:agent}
Recent work reframes RAG as an agent-based pipeline. Surveys on agent-based RAG describe systems in which LLM-based agents actively control query rewriting, retrieval, and answer generation through planning, tool use, reflection, and multi-agent coordination, rather than following a static single-pass workflow~\citep{singh2025agentic}. In parallel, personalization studies show a shift from personalized RAG, which injects user priors into retrieval and generation stages, to personalized agents that maintain user models and adapt retrieval strategies over time~\citep{li2025survey}. From this perspective, agents serve as controllers of the RAG process, contextualizing retrieval and selecting evidence under user- and task-specific constraints.

\paragraph{Agent-based Textual RAG.}
Concrete architectures realize this idea by decomposing the RAG pipeline into interacting agents with specialized roles. MAIN-RAG~\citep{chang2025main} coordinates predictor, judge, and final predictor agents to filter noisy documents via consensus scoring and adaptive thresholds, yielding training-free gains in accuracy and faithfulness. MA-RAG~\citep{nguyen2025ma} further separates planning, step definition, evidence extraction, and QA into distinct chain-of-thought agents, improving multi-hop and ambiguous QA without fine-tuning. MMOA-RAG~\citep{chen2025improving} adopts an optimization view by modeling each RAG component as a cooperative RL agent under a shared task-level reward, aligning local decisions with end-to-end QA performance. AU-RAG~\citep{jang2024rag} extends this paradigm by using an agent to select and query heterogeneous, frequently updated data sources through descriptive metadata rather than fixed vector indices, enabling more flexible retrieval across APIs and disparate stores. Together, these methods characterize agent-based RAG as a modular and goal-driven paradigm, where specialized agents are coordinated under explicit global objectives to improve robustness, adaptability, and end-to-end performance. For multimodal document understanding~\citep{abootorabi2025ask}, this paradigm naturally extends to settings in which agents allocate queries across text, images, tables, graphs, and web sources, maintain cross-modal state over long interactions, and evaluate correctness using task-aligned multimodal signals.

\paragraph{Agent-based Multimodal RAG.}
Agent-based multimodal RAG instantiates these patterns by deploying agents that coordinate retrieval and generation across modalities. Agents dynamically formulate sub-queries, select retrieval strategies, and fuse evidence from text, images, tables, and layout blocks according to task requirements (see Figure~\ref{fig:graph_agent} (b)). Through multi-agent collaboration, systems can perform iterative reasoning, verification, and evidence refinement, which improves both accuracy and transparency. ViDoRAG~\citep{wang2025vidorag} follows an iterative workflow in which exploration, summarization, and reflection agents traverse visually rich corpora to progressively refine retrieval results and answers. HM-RAG~\citep{liu2025hm}, in contrast, adopts a more structured organization, combining a Decomposition Agent for query rewriting, modality-specific Retrieval Agents for parallel evidence collection, and a Decision Agent that integrates outputs through consistency voting. Patho-AgenticRAG~\citep{zhang2025patho} extends this paradigm to the medical domain by coupling task decomposition and search agents with reinforcement-learned policies, enabling robust joint text and image retrieval while reducing hallucinations in diagnostic reasoning.

Other multimodal frameworks further expand the design space of agent roles. HEAR~\citep{chen2025hear} tightly couples VLM-based document parsing with a closed-loop multi-agent reasoning process, re-invoking parsers when cross-modal inconsistencies are detected. SLEUTH~\citep{liu2025resolving} adopts a coarse-to-fine agent scheme that filters and distills salient textual and visual evidence into compact contexts for long-document understanding. Overall, agent-based multimodal RAG reframes multimodal retrieval and reasoning as a coordinated process among specialized agents for query formulation, modality allocation, and evidence validation. By enabling adaptive retrieval depth and structured cross-modal reasoning, it moves beyond static retrieve-then-read pipelines and is well suited for complex multimodal documents and domain-specific tasks.

\paragraph{Discussion and open challenges.}
Despite their flexibility, agent-based multimodal RAG systems introduce substantial computational and economic overhead. Multi-agent coordination often requires repeated LLM calls for planning, decomposition, retrieval, verification, and reflection, which can significantly increase latency and inference cost compared to single-pass RAG pipelines~\citep{singh2025agentic,li2025survey}. This issue is exacerbated in multimodal settings, where agents may invoke expensive vision-language models, document parsers, or external tools multiple times. Balancing performance gains with practical efficiency thus remains a key challenge. Promising directions include adaptive agent activation, where agents are invoked conditionally based on task complexity or uncertainty, lightweight proxy models for early-stage filtering, and shared memory or caching mechanisms to reduce redundant reasoning and retrieval~\citep{chang2025main,liu2025resolving}.

A second open challenge concerns coordination and optimization in increasingly complex agent ecosystems. As the number of agents and modalities grows, designing stable interaction protocols, credit assignment mechanisms, and global objectives becomes nontrivial, and poorly aligned agents may amplify noise or propagate errors across modalities~\citep{chen2025improving,wang2025vidorag}. Future research may benefit from tighter integration of learning-based controllers, such as reinforcement learning or meta-learning, to automatically discover effective agent roles, communication patterns, and stopping criteria under resource constraints~\citep{chen2025improving,zhang2025patho}. More generally, principled evaluation frameworks that jointly measure answer quality, faithfulness, interpretability, and cost will be critical for guiding the development of scalable and reliable agent-based multimodal RAG systems in real-world deployments.

%% file: table/dataset.tex
\begin{table*}[ht]
\centering
\resizebox{\linewidth}{!}{
\begin{tabular}{l|p{13cm}}
\toprule
\rowcolor{gray!25} \textbf{Dataset} & \textbf{Features} \\
\midrule
PlotQA~\citep{methani2020plotqa} & Bridges the gap to real-world plots with a large-scale dataset built from authentic charts and crowd-sourced questions, requiring complex reasoning and out-of-vocabulary answers beyond fixed vocabularies. \\
\rowcolor{gray!25} TabFQuAD~\citep{d2020fquad} & Evaluates TableQA models in realistic industry settings using a French table question-answering dataset enhanced with GPT-4V generated queries. \\
DocVQA~\citep{mathew2021docvqa} & Highlights the gap between human and model performance on structured document understanding using a large-scale dataset from UCSF Industry collections. \\
\rowcolor{gray!25} VisualMRC~\cite{tanaka2021visualmrc} &  Builds a visual machine reading comprehension dataset from multi-domain webpage documents to advance natural language understanding and generation from document images. \\
ChartQA~\citep{masry2022chartqa} & Constructs a large-scale chart QA benchmark with human-written and generated questions to evaluate models on complex logical, arithmetic, and visual reasoning over charts. \\
\rowcolor{gray!25} InfoVQA~\citep{mathew2022infographicvqa} & Benchmarks models on reasoning over layout, text, and visuals using a diverse infographic QA dataset highlighting the human–machine gap. \\
TAT-DQA~\citep{zhu2022towards} &  Samples financial reports with semi-structured tables and text to build a document QA dataset requiring discrete numerical reasoning, highlighting the gap between models and human experts. \\
\rowcolor{gray!25} ScienceQA~\citep{saikh2022scienceqa} & Introduces a multimodal benchmark of diverse science questions with annotated answers, lectures, and explanations to evaluate and enhance models’ reasoning through chain-of-thought. \\
DUDE~\citep{van2023document} &  Creates a practical benchmark from multi-industry, multi-domain visually-rich documents to evaluate document AI on real-world, multi-task, and low-resource scenarios. \\
\rowcolor{gray!25} SlideVQA~\citep{tanaka2023slidevqa} &  Builds a multi-image document QA dataset from slide decks to enable complex single-hop, multi-hop, and numerical reasoning, highlighting the gap between models and human performance. \\
ArXivQA~\citep{li2024multimodal} & Builds a scientific QA dataset from ArXiv papers to boost LVLMs’ ability in interpreting abstract figures and improving mathematical reasoning. \\
\rowcolor{gray!25} MMLongBench-Doc~\citep{ma2024mmlongbench} &  Constructs a long-context multimodal benchmark from lengthy PDFs with cross-page questions to evaluate LVLMs on document understanding. \\
PaperTab~\citep{hui2024uda} & Extracts academic papers in PDF format for extractive, yes/no, and free-form QA. \\
\rowcolor{gray!25} FetaTab~\citep{hui2024uda} & Gathers world knowledge documents in PDF and HTML format for free-form QA. \\
SPIQA~\citep{pramanick2024spiqa} & Creates a large-scale QA dataset from scientific papers that integrates text with complex figures and tables to evaluate and advance multimodal understanding in research articles. \\
\rowcolor{gray!25} LongDocUrl~\citep{deng2024longdocurl} & Integrates long-document understanding, numerical reasoning, and cross-element locating into a large-scale benchmark to expose critical gaps in current LVLMs. \\
\midrule
ViDoRe~\citep{faysse2024colpali} & Unifies academic tasks with diverse document types and practical tasks across multiple domains and languages to comprehensively evaluate multimodal document retrieval. \\
\rowcolor{gray!25} VisR-Bench~\citep{chen2024sv} & Selects diverse visually-rich documents with tables, charts, and diagrams, and generate verified QA pairs using GPT-4o to create a benchmark highlighting multimodal reasoning and quality assurance. \\
M3DoCVQA~\citep{cho2024m3docrag} & Evaluates open-domain DocVQA with M3DoCVQA, a large multi-page PDF benchmark requiring multi-hop, multimodal reasoning across text and visual elements. \\
\rowcolor{gray!25} VisDoMBench~\citep{suri2025visdom} &  Leverages multiple documents with diverse modalities such as tables, charts, and slides, requiring cross-document reasoning, modality fusion, and verifiable answers. \\
ViDoSeek~\citep{wang2025vidorag} &  Unifies queries and large corpora of visually rich documents to enable complex reasoning beyond image-based QA, emphasizing multimodal retrieval, cross-document comprehension, and unique answer generation. \\
\rowcolor{gray!25} OpenDocVQA~\citep{tanaka2025vdocrag} &  Combines diverse document types, formats, and modalities into a unified open-domain collection to train and evaluate retrieval and QA models on visually-rich documents. \\
UniDoc-Bench~\citep{peng2025unidoc} & Provides a unified, large-scale benchmark for evaluating multimodal RAG on real-world documents, enabling fair comparison across text-only, image-only, and multimodal retrieval settings. \\
\rowcolor{gray!25} BBox-DocVQA~\citep{yu2025bbox} & Introduces a bounding-box–grounded DocVQA benchmark to evaluate fine-grained spatial grounding and reasoning in visually-rich documents. \\
\bottomrule
\end{tabular}
}
\caption{Popular datasets and benchmarks in multimodal RAG for document understanding, along with detailed descriptions of their data sources and characteristics.}
\label{tab:appendix_db}
\end{table*}


%% file: table/key_contribution.tex
\begin{table*}[ht]
\centering
\resizebox{\linewidth}{!}{
\begin{tabular}{l|p{14cm}}
\toprule
\rowcolor{gray!25} \textbf{Method} & \textbf{Key Contribution Summary} \\
\midrule
DSE~\citep{ma2024unifying} & Encodes document screenshots with VLMs for retrieval, avoiding parsing and preserving full multimodal information. \\ 
\rowcolor{gray!25} ColPali~\citep{faysse2024colpali} & Embeds document page images into multi-vector representations with late interaction matching for efficient end-to-end retrieval. \\
ColQwen2~\citep{faysse2024colpali} & Extends Qwen2-VL-2B to generate ColBERT-style multi-vector representations for complex text–image tasks, similar to ColPali. \\
\rowcolor{gray!25} CREAM~\citep{zhang2024cream} & Combines coarse-to-fine retrieval with multi-page visual attention pooling, enabling effective integration of multimodal document information. \\
VisRAG~\citep{yu2024visrag} & Introduces a VLM-based RAG pipeline that embeds documents as images, preserving multimodal information and avoiding text-parsing loss. \\
\rowcolor{gray!25} SV-RAG~\citep{chen2024sv} & Introduces a framework where MLLMs act as retriever and generator with two adapters for retrieval and question answering. \\
M3DocRAG~\citep{cho2024m3docrag} & Unifies retrieval and reasoning across text, charts, and figures, enabling flexible multi-hop DocVQA over single or multi-page documents. \\
\rowcolor{gray!25} VisDoMRAG~\citep{suri2025visdom} & Introduces consistency-constrained modality fusion for unified multi-step reasoning across visual and textual modalities in multimodal document QA. \\
GME~\citep{zhang2025bridging} & Advances universal multimodal retrieval by leveraging a synthetic fused-modal training dataset and an MLLM-based dense retriever, achieving state-of-the-art performance on the new UMR Benchmark. \\
\rowcolor{gray!25} ViDoRAG~\citep{wang2025vidorag} & Leverages a multi-agent, Gaussian Mixture Model-based hybrid retrieval and iterative reasoning workflow for complex understanding of visually rich documents. \\
HM-RAG~\citep{liu2025hm} &  Decomposes queries hierarchically, retrieves from diverse modalities, and integrates results via consistency voting for robust multimodal reasoning. \\
\rowcolor{gray!25} VDocRAG~\citep{tanaka2025vdocrag} & Unifies visually-rich documents into image-based representations and design self-supervised pre-training tasks that compress visual information into dense tokens aligned with textual content for retrieval-augmented generation. \\
FRAG~\citep{huang2025frag} & Selects relevant frames to improve multimodal model generation efficiency and performance. \\
\rowcolor{gray!25} MG-RAG~\citep{xu2025multi} & Integrates hierarchical encoding, modality-aware retrieval, and VLM-based candidate filtering to effectively handle visually-rich documents. \\
VRAG-RL~\citep{wang2025vrag} & Introduces an RL framework that enables VLMs to reason effectively over documents from pages to fine-grained regions. \\
\rowcolor{gray!25} CoRe-MMRAG~\citep{tian2025core} & Reconciles inconsistencies between parametric and retrieved multimodal knowledge through a four-stage framework with specialized training for reliable answer generation. \\
Light-ColPali~\citep{ma2025towards} &  Reduces memory usage in Visualized Document Retrieval by applying optimized token merging, preserving over 94\% effectiveness with as little as 2.8\% of the original memory. \\
\rowcolor{gray!25} MM-R5~\citep{xu2025mm} & Enhances multimodal document retrieval by integrating supervised fine-tuning and reinforcement learning with reasoning chains and task-specific rewards. \\
SimpleDoc~\citep{jain2025simpledoc} & Combines embedding-based retrieval with summary-based re-ranking, enabling efficient multi-page reasoning with a single VLM agent. \\
\rowcolor{gray!25} VisChunk~\citep{tripathi2025vision}& Leverages multimodal cues to chunk documents while preserving structural and semantic coherence, enhancing downstream RAG performance. \\
DocVQA-RAP~\citep{yu2025beyond} & Proposes a utility-driven retrieval method for VDQA that scores evidence by its predicted contribution to answer quality, reducing reliance on mere semantic relevance. \\
\rowcolor{gray!25} RL-QR~\citep{cha2025generalized} & Applies reinforcement learning–based query rewriting without annotations, tailoring rewriters to specific retrievers and boosting RAG performance across text and multimodal databases. \\
MMRAG-DocQA~\citep{gong2025mmrag} & Leverages hierarchical indexing and multi-granularity retrieval to connect in-page and cross-page multimodal evidence, enabling accurate reasoning over long, modality-rich documents. \\
\rowcolor{gray!25} Patho-AgenticRAG~\citep{zhang2025patho} & Enables joint text–image retrieval from pathology textbooks with agentic reasoning and multi-turn search, reducing hallucinations and improving diagnostic accuracy. \\
M2IO-R1~\citep{xiao2025m2io} & Enables multimodal inputs and outputs in RAG with an RL-based framework using an Inserter module for controllable image selection and placement. \\
\rowcolor{gray!25} mKG-RAG~\citep{yuan2025mkg} & Enhances RAG-based VQA by constructing multimodal knowledge graphs and employing dual-stage, question-aware retrieval to provide structured, modality-aligned knowledge for more accurate generation. \\
DB3Team-RAG~\citep{xia2025db3} & Integrates domain-specific multimodal retrieval pipelines with unified LLM tuning and refusal training. \\
\rowcolor{gray!25} PREMIR~\citep{choi2025zero} & Boosts multimodal retrieval by generating cross-modal pre-questions, enabling robust token-level matching across domains and languages. \\
ReDocRAG~\citep{lopez2025enhancing} & Enhances Document VQA by retrieving and reranking key evidence, achieving higher accuracy on multi-page datasets with reduced memory demand. \\
\rowcolor{gray!25} CMRAG~\citep{chen2025cmrag} & Leverages co-modality representations of text and images for joint retrieval and generation, enabling more effective document visual question answering than text-only or vision-only RAG methods. \\

\bottomrule
\end{tabular}
}
\caption{Key contributions of multimodal RAG methods for document understanding (Part1).}
\label{tab:appendix_key_contribution_part1}
\end{table*}

\begin{table*}[ht]
\centering
\resizebox{\linewidth}{!}{
\begin{tabular}{l|p{14cm}}
\toprule
\rowcolor{gray!25} \textbf{Method} & \textbf{Key Contribution Summary} \\
\midrule
MoLoRAG~\citep{wu2025molorag} & Enhances multi-modal, multi-page DocQA by combining semantic and logic-aware retrieval through page-graph traversal, enabling LVLMs to capture overlooked logical connections for more accurate answers. \\
\rowcolor{gray!25} SERVAL~\citep{nguyen2025serval} & Leverages vision–language models to generate textual descriptions of document images and embed them with a text encoder for scalable zero-shot visual document retrieval. \\
MetaEmbed~\citep{xiao2025metaembed} & Employs learnable Meta Tokens to generate compact multi-vector embeddings, enabling scalable test-time trade-offs between retrieval quality and efficiency. \\
\rowcolor{gray!25} DocPruner~\citep{yan2025docpruner} & Adaptively prunes redundant patch-level embeddings based on intra-document attention, substantially reducing storage costs for multi-vector VDR while preserving retrieval effectiveness. \\
RECON~\citep{wangrecon} & Proposes a two-stage multimodal knowledge graph construction framework for visually rich documents, featuring intra-page reflection to extract textual–visual entity relations and inter-page connection to integrate cross-page multimodal relations into a global graph. \\
\rowcolor{gray!25} LAD-RAG~\citep{sourati2025lad} & Proposes a layout-aware dynamic RAG framework that constructs a symbolic document graph to model layout structure and cross-page dependencies, and enables adaptive evidence retrieval through LLM-guided interaction with neural and symbolic indices.\\
HEAVEN~\citep{kim2025hybrid} & Proposes a two-stage hybrid-vector retrieval framework that combines single-vector candidate retrieval over visually summarized pages with efficient multi-vector reranking for visually rich documents. \\
\rowcolor{gray!25} DREAM~\citep{zhang2025dream} & Proposes a retrieval-enhanced multimodal framework that combines confidence-based and embedding-based document retrieval with a decoupled cross-page attention-aware MLLM to enable effective multi-page document understanding and visual question answering.\\
MARA~\citep{wu2025mara} & Proposes a multimodal adaptive RAG framework that introduces query-aligned document representations for retrieval and a self-reflective evidence controller to dynamically incorporate sufficient multimodal evidence during generation. \\
\rowcolor{gray!25} HEAR~\citep{chen2025hear} & Introduces a holistic extraction and agentic reasoning framework that tightly couples VLM-based structured document parsing with a closed-loop, multi-agent cross-modal reasoning system, enabling active verification and conflict-driven re-engagement for complex multimodal document understanding. \\
HPC-ColPali~\citep{bach2025hierarchical} & Proposes a hierarchical patch compression framework that improves the efficiency of multi-vector document retrieval through quantization and attention-guided pruning while maintaining retrieval accuracy. \\
\rowcolor{gray!25} RegionRAG~\citep{li2025regionrag} & Proposes a region-level multimodal RAG framework that identifies and retrieves query-relevant visual regions via hybrid supervision and dynamic region grouping, reducing redundant visual context while improving retrieval and generation accuracy. \\
IndustryRAG~\citep{lim2025distilling} & Proposes an efficient knowledge distillation framework that transfers complementary domain and visual–structural knowledge from LLMs and VLMs into a compact domain-specific retriever, enabling effective RAG for industrial documents with complex structural elements. \\
\rowcolor{gray!25} COLMATE~\citep{masry2025colmate} & Proposes a multimodal document retrieval model with OCR-aware pretraining and late-interaction scoring, better aligning representation learning with multimodal document retrieval. \\
LILaC~\citep{yun2025lilac} & Proposes a multimodal retrieval framework that models documents with a layered component graph and performs late interaction–based subgraph retrieval, enabling efficient multi-granular retrieval and effective multihop reasoning across multimodal components. \\
\rowcolor{gray!25} HKRAG~\citep{tong2025hkrag} & Proposes a holistic multimodal RAG framework that jointly models salient and fine-print knowledge through hybrid masking–based retrieval and an uncertainty-guided agentic generator, enabling more complete and accurate understanding of visually rich documents. \\
SLEUTH~\citep{liu2025resolving} & Proposes a multi-agent, coarse-to-fine framework that collaboratively filters and distills salient textual and visual evidence from retrieved pages, synthesizing an evidence-dense multimodal context for effective long-document understanding. \\
\rowcolor{gray!25} Snappy~\citep{georgiou2025spatially} & Proposes a hybrid multimodal retrieval framework that fuses ColPali’s patch-level similarity with OCR-extracted regions via spatial relevance mapping, enabling precise region-level evidence selection for RAG without additional training. \\

\bottomrule
\end{tabular}
}
\caption{Key contributions of multimodal RAG methods for document understanding (Part2).}
\label{tab:appendix_key_contribution_part2}
\end{table*}

%% file: appendix/key_contribution.tex
\section{Key Contribution Summary}
\label{sec:appendix_key_contribution}
Table~\ref{tab:appendix_key_contribution_part1} and ~\ref{tab:appendix_key_contribution_part2} presents a consolidated overview of the key contributions of existing multimodal RAG approaches for document understanding. By systematically organizing and comparing these methods, this survey highlights the breadth of design choices and research directions in the field. Such a structured summary not only helps researchers quickly grasp the state of the art, but also clarifies common trends, complementary strengths, and open challenges. In doing so, it serves as a reference point for guiding future work and motivating new directions in multimodal retrieval and reasoning for complex document understanding.

%% file: main.bib
@article{subramani2020survey,
  title={A survey of deep learning approaches for ocr and document understanding},
  author={Subramani, Nishant and Matton, Alexandre and Greaves, Malcolm and Lam, Adrian},
  journal={arXiv preprint arXiv:2011.13534},
  year={2020}
}

@article{ding2024deep,
  title={Deep learning based visually rich document content understanding: A survey},
  author={Ding, Yihao and Han, Soyeon Caren and Lee, Jean and Hovy, Eduard},
  journal={arXiv preprint arXiv:2408.01287},
  year={2024}
}

@article{gu2021unidoc,
  title={Unidoc: Unified pretraining framework for document understanding},
  author={Gu, Jiuxiang and Kuen, Jason and Morariu, Vlad I and Zhao, Handong and Jain, Rajiv and Barmpalios, Nikolaos and Nenkova, Ani and Sun, Tong},
  journal={Advances in Neural Information Processing Systems},
  volume={34},
  pages={39--50},
  year={2021}
}

@inproceedings{appalaraju2021docformer,
  title={Docformer: End-to-end transformer for document understanding},
  author={Appalaraju, Srikar and Jasani, Bhavan and Kota, Bhargava Urala and Xie, Yusheng and Manmatha, R},
  booktitle={Proceedings of the IEEE/CVF international conference on computer vision},
  pages={993--1003},
  year={2021}
}

@article{shi2016end,
  title={An end-to-end trainable neural network for image-based sequence recognition and its application to scene text recognition},
  author={Shi, Baoguang and Bai, Xiang and Yao, Cong},
  journal={IEEE transactions on pattern analysis and machine intelligence},
  volume={39},
  number={11},
  pages={2298--2304},
  year={2016},
  publisher={IEEE}
}

@inproceedings{park2019cord,
  title={Cord: a consolidated receipt dataset for post-ocr parsing},
  author={Park, Seunghyun and Shin, Seung and Lee, Bado and Lee, Junyeop and Surh, Jaeheung and Seo, Minjoon and Lee, Hwalsuk},
  booktitle={Workshop on Document Intelligence at NeurIPS 2019},
  year={2019}
}

@article{ding2025vrd,
  title={VRD-IU: Lessons from Visually Rich Document Intelligence and Understanding},
  author={Ding, Yihao and Han, Soyeon Caren and Li, Yan and Poon, Josiah},
  journal={arXiv preprint arXiv:2506.01388},
  year={2025}
}

@inproceedings{duan2025docopilot,
  title={Docopilot: Improving Multimodal Models for Document-Level Understanding},
  author={Duan, Yuchen and Chen, Zhe and Hu, Yusong and Wang, Weiyun and Ye, Shenglong and Shi, Botian and Lu, Lewei and Hou, Qibin and Lu, Tong and Li, Hongsheng and others},
  booktitle={Proceedings of the Computer Vision and Pattern Recognition Conference},
  pages={4026--4037},
  year={2025}
}

@article{xiong2025docr1,
  title={DocR1: Evidence Page-Guided GRPO for Multi-Page Document Understanding},
  author={Xiong, Junyu and Wang, Yonghui and Zhao, Weichao and Liu, Chenyu and Yin, Bing and Zhou, Wengang and Li, Houqiang},
  journal={arXiv preprint arXiv:2508.07313},
  year={2025}
}

@article{yu2025docthinker,
  title={DocThinker: Explainable Multimodal Large Language Models with Rule-based Reinforcement Learning for Document Understanding},
  author={Yu, Wenwen and Yang, Zhibo and Liu, Yuliang and Bai, Xiang},
  journal={arXiv preprint arXiv:2508.08589},
  year={2025}
}

@article{zhou2024doge,
  title={Doge: Towards versatile visual document grounding and referring},
  author={Zhou, Yinan and Chen, Yuxin and Lin, Haokun and Yang, Shuyu and Zhu, Li and Qi, Zhongang and Ma, Chen and Shan, Ying},
  journal={arXiv preprint arXiv:2411.17125},
  year={2024}
}

@article{nassar2025smoldocling,
  title={SmolDocling: An ultra-compact vision-language model for end-to-end multi-modal document conversion},
  author={Nassar, Ahmed and Marafioti, Andres and Omenetti, Matteo and Lysak, Maksym and Livathinos, Nikolaos and Auer, Christoph and Morin, Lucas and de Lima, Rafael Teixeira and Kim, Yusik and Gurbuz, A Said and others},
  journal={arXiv preprint arXiv:2503.11576},
  year={2025}
}

@article{ye2023mplug,
  title={mplug-docowl: Modularized multimodal large language model for document understanding},
  author={Ye, Jiabo and Hu, Anwen and Xu, Haiyang and Ye, Qinghao and Yan, Ming and Dan, Yuhao and Zhao, Chenlin and Xu, Guohai and Li, Chenliang and Tian, Junfeng and others},
  journal={arXiv preprint arXiv:2307.02499},
  year={2023}
}

@article{hu2024mplug,
  title={mplug-docowl 1.5: Unified structure learning for ocr-free document understanding},
  author={Hu, Anwen and Xu, Haiyang and Ye, Jiabo and Yan, Ming and Zhang, Liang and Zhang, Bo and Li, Chen and Zhang, Ji and Jin, Qin and Huang, Fei and others},
  journal={arXiv preprint arXiv:2403.12895},
  year={2024}
}

@article{hu2024mplug2,
  title={mplug-docowl2: High-resolution compressing for ocr-free multi-page document understanding},
  author={Hu, Anwen and Xu, Haiyang and Zhang, Liang and Ye, Jiabo and Yan, Ming and Zhang, Ji and Jin, Qin and Huang, Fei and Zhou, Jingren},
  journal={arXiv preprint arXiv:2409.03420},
  year={2024}
}

@article{deng2024longdocurl,
  title={Longdocurl: a comprehensive multimodal long document benchmark integrating understanding, reasoning, and locating},
  author={Deng, Chao and Yuan, Jiale and Bu, Pi and Wang, Peijie and Li, Zhong-Zhi and Xu, Jian and Li, Xiao-Hui and Gao, Yuan and Song, Jun and Zheng, Bo and others},
  journal={arXiv preprint arXiv:2412.18424},
  year={2024}
}

@article{ma2024mmlongbench,
  title={Mmlongbench-doc: Benchmarking long-context document understanding with visualizations},
  author={Ma, Yubo and Zang, Yuhang and Chen, Liangyu and Chen, Meiqi and Jiao, Yizhu and Li, Xinze and Lu, Xinyuan and Liu, Ziyu and Ma, Yan and Dong, Xiaoyi and others},
  journal={Advances in Neural Information Processing Systems},
  volume={37},
  pages={95963--96010},
  year={2024}
}

@article{liu2025hm,
  title={Hm-rag: Hierarchical multi-agent multimodal retrieval augmented generation},
  author={Liu, Pei and Liu, Xin and Yao, Ruoyu and Liu, Junming and Meng, Siyuan and Wang, Ding and Ma, Jun},
  journal={arXiv preprint arXiv:2504.12330},
  year={2025}
}

@article{han2025mdocagent,
  title={Mdocagent: A multi-modal multi-agent framework for document understanding},
  author={Han, Siwei and Xia, Peng and Zhang, Ruiyi and Sun, Tong and Li, Yun and Zhu, Hongtu and Yao, Huaxiu},
  journal={arXiv preprint arXiv:2503.13964},
  year={2025}
}

@article{wang2025multi,
  title={Multi-Agent Interactive Question Generation Framework for Long Document Understanding},
  author={Wang, Kesen and Toibazar, Daulet and Alfulayt, Abdulrahman and Albadawi, Abdulaziz S and Alkahtani, Ranya A and Ibrahim, Asma A and Alhomoud, Haneen A and Mohamed, Sherif and Moreno, Pedro J},
  journal={arXiv preprint arXiv:2507.20145},
  year={2025}
}

@article{wu2150tabagent,
  title={TabAgent: A Multi-Agent Table Extraction Framework for Unstructured Documents},
  author={Wu, Jingfei and Shen, Chaoyuan and Deng, Qiyan and Wang, Yuping and Li, Jiajun and Deng, Yuhao and Yu, Minghe},
  journal={Proceedings of the VLDB Endowment. ISSN},
  volume={2150},
  pages={8097},
  yeal={2025}
}

@article{yu2025visual,
  title={Visual Document Understanding and Question Answering: A Multi-Agent Collaboration Framework with Test-Time Scaling},
  author={Yu, Xinlei and Chen, Zhangquan and Zhang, Yudong and Lu, Shilin and Shen, Ruolin and Zhang, Jiangning and Hu, Xiaobin and Fu, Yanwei and Yan, Shuicheng},
  journal={arXiv preprint arXiv:2508.03404},
  year={2025}
}

@article{arslan2024survey,
  title={A Survey on RAG with LLMs},
  author={Arslan, Muhammad and Ghanem, Hussam and Munawar, Saba and Cruz, Christophe},
  journal={Procedia computer science},
  volume={246},
  pages={3781--3790},
  year={2024},
  publisher={Elsevier}
}

@inproceedings{fan2024survey,
  title={A survey on rag meeting llms: Towards retrieval-augmented large language models},
  author={Fan, Wenqi and Ding, Yujuan and Ning, Liangbo and Wang, Shijie and Li, Hengyun and Yin, Dawei and Chua, Tat-Seng and Li, Qing},
  booktitle={Proceedings of the 30th ACM SIGKDD conference on knowledge discovery and data mining},
  pages={6491--6501},
  year={2024}
}

@article{wang2022text,
  title={Text embeddings by weakly-supervised contrastive pre-training},
  author={Wang, Liang and Yang, Nan and Huang, Xiaolong and Jiao, Binxing and Yang, Linjun and Jiang, Daxin and Majumder, Rangan and Wei, Furu},
  journal={arXiv preprint arXiv:2212.03533},
  year={2022}
}

@article{li2023towards,
  title={Towards general text embeddings with multi-stage contrastive learning},
  author={Li, Zehan and Zhang, Xin and Zhang, Yanzhao and Long, Dingkun and Xie, Pengjun and Zhang, Meishan},
  journal={arXiv preprint arXiv:2308.03281},
  year={2023}
}

@article{chen2024bge,
  title={Bge m3-embedding: Multi-lingual, multi-functionality, multi-granularity text embeddings through self-knowledge distillation},
  author={Chen, Jianlv and Xiao, Shitao and Zhang, Peitian and Luo, Kun and Lian, Defu and Liu, Zheng},
  journal={arXiv preprint arXiv:2402.03216},
  year={2024}
}

@inproceedings{khattab2020colbert,
  title={Colbert: Efficient and effective passage search via contextualized late interaction over bert},
  author={Khattab, Omar and Zaharia, Matei},
  booktitle={Proceedings of the 43rd International ACM SIGIR conference on research and development in Information Retrieval},
  pages={39--48},
  year={2020}
}

@article{abootorabi2025ask,
  title={Ask in any modality: A comprehensive survey on multimodal retrieval-augmented generation},
  author={Abootorabi, Mohammad Mahdi and Zobeiri, Amirhosein and Dehghani, Mahdi and Mohammadkhani, Mohammadali and Mohammadi, Bardia and Ghahroodi, Omid and Baghshah, Mahdieh Soleymani and Asgari, Ehsaneddin},
  journal={arXiv preprint arXiv:2502.08826},
  year={2025}
}

@article{mei2025survey,
  title={A survey of multimodal retrieval-augmented generation},
  author={Mei, Lang and Mo, Siyu and Yang, Zhihan and Chen, Chong},
  journal={arXiv preprint arXiv:2504.08748},
  year={2025}
}

@article{faysse2024colpali,
  title={Colpali: Efficient document retrieval with vision language models},
  author={Faysse, Manuel and Sibille, Hugues and Wu, Tony and Omrani, Bilel and Viaud, Gautier and Hudelot, C{\'e}line and Colombo, Pierre},
  journal={arXiv preprint arXiv:2407.01449},
  year={2024}
}

@article{yu2024visrag,
  title={Visrag: Vision-based retrieval-augmented generation on multi-modality documents},
  author={Yu, Shi and Tang, Chaoyue and Xu, Bokai and Cui, Junbo and Ran, Junhao and Yan, Yukun and Liu, Zhenghao and Wang, Shuo and Han, Xu and Liu, Zhiyuan and others},
  journal={arXiv preprint arXiv:2410.10594},
  year={2024}
}

@article{wang2025vrag,
  title={VRAG-RL: Empower Vision-Perception-Based RAG for Visually Rich Information Understanding via Iterative Reasoning with Reinforcement Learning},
  author={Wang, Qiuchen and Ding, Ruixue and Zeng, Yu and Chen, Zehui and Chen, Lin and Wang, Shihang and Xie, Pengjun and Huang, Fei and Zhao, Feng},
  journal={arXiv preprint arXiv:2505.22019},
  year={2025}
}

@article{choi2025zero,
  title={Zero-shot Multimodal Document Retrieval via Cross-modal Question Generation},
  author={Choi, Yejin and Park, Jaewoo and Yoon, Janghan and Kim, Saejin and Jeon, Jaehyun and Yu, Youngjae},
  journal={arXiv preprint arXiv:2508.17079},
  year={2025}
}

@article{chen2024sv,
  title={SV-RAG: LoRA-Contextualizing Adaptation of MLLMs for Long Document Understanding},
  author={Chen, Jian and Zhang, Ruiyi and Zhou, Yufan and Yu, Tong and Dernoncourt, Franck and Gu, Jiuxiang and Rossi, Ryan A and Chen, Changyou and Sun, Tong},
  journal={arXiv preprint arXiv:2411.01106},
  year={2024}
}

@article{cho2024m3docrag,
  title={M3docrag: Multi-modal retrieval is what you need for multi-page multi-document understanding},
  author={Cho, Jaemin and Mahata, Debanjan and Irsoy, Ozan and He, Yujie and Bansal, Mohit},
  journal={arXiv preprint arXiv:2411.04952},
  year={2024}
}

@article{xiao2025m2io,
  title={M2IO-R1: An Efficient RL-Enhanced Reasoning Framework for Multimodal Retrieval Augmented Multimodal Generation},
  author={Xiao, Zhiyou and Yu, Qinhan and Li, Binghui and Chen, Geng and Chen, Chong and Zhang, Wentao},
  journal={arXiv preprint arXiv:2508.06328},
  year={2025}
}

@article{gong2025mmrag,
  title={MMRAG-DocQA: A Multi-Modal Retrieval-Augmented Generation Method for Document Question-Answering with Hierarchical Index and Multi-Granularity Retrieval},
  author={Gong, Ziyu and Huang, Yihua and Mai, Chengcheng},
  journal={arXiv preprint arXiv:2508.00579},
  year={2025}
}

@article{tripathi2025vision,
  title={Vision-Guided Chunking Is All You Need: Enhancing RAG with Multimodal Document Understanding},
  author={Tripathi, Vishesh and Odapally, Tanmay and Das, Indraneel and Allu, Uday and Ahmed, Biddwan},
  journal={arXiv preprint arXiv:2506.16035},
  year={2025}
}

@article{tian2025core,
  title={CoRe-MMRAG: Cross-Source Knowledge Reconciliation for Multimodal RAG},
  author={Tian, Yang and Liu, Fan and Zhang, Jingyuan and Hu, Yupeng and Nie, Liqiang and others},
  journal={arXiv preprint arXiv:2506.02544},
  year={2025}
}

@article{ma2024unifying,
  title={Unifying multimodal retrieval via document screenshot embedding},
  author={Ma, Xueguang and Lin, Sheng-Chieh and Li, Minghan and Chen, Wenhu and Lin, Jimmy},
  journal={arXiv preprint arXiv:2406.11251},
  year={2024}
}

@inproceedings{zhang2024cream,
  title={CREAM: coarse-to-fine retrieval and multi-modal efficient tuning for document VQA},
  author={Zhang, Jinxu and Yu, Yongqi and Zhang, Yu},
  booktitle={Proceedings of the 32nd ACM International Conference on Multimedia},
  pages={925--934},
  year={2024}
}

@article{yuan2025mkg,
  title={mKG-RAG: Multimodal Knowledge Graph-Enhanced RAG for Visual Question Answering},
  author={Yuan, Xu and Ning, Liangbo and Fan, Wenqi and Li, Qing},
  journal={arXiv preprint arXiv:2508.05318},
  year={2025}
}

@article{jain2025simpledoc,
  title={SimpleDoc: Multi-Modal Document Understanding with Dual-Cue Page Retrieval and Iterative Refinement},
  author={Jain, Chelsi and Wu, Yiran and Zeng, Yifan and Liu, Jiale and Shao, Zhenwen and Wu, Qingyun and Wang, Huazheng and others},
  journal={arXiv preprint arXiv:2506.14035},
  year={2025}
}

@article{chen2025cmrag,
  title={CMRAG: Co-modality-based document retrieval and visual question answering},
  author={Chen, Wang and Qi, Guanqiang and Li, Weikang and Li, Yang},
  journal={arXiv preprint arXiv:2509.02123},
  year={2025}
}

@article{wu2025molorag,
  title={MoLoRAG: Bootstrapping Document Understanding via Multi-modal Logic-aware Retrieval},
  author={Wu, Xixi and Tan, Yanchao and Hou, Nan and Zhang, Ruiyang and Cheng, Hong},
  journal={arXiv preprint arXiv:2509.07666},
  year={2025}
}

@article{xu2025multi,
  title={A Multi-Granularity Retrieval Framework for Visually-Rich Documents},
  author={Xu, Mingjun and Wang, Zehui and Cai, Hengxing and Zhong, Renxin},
  journal={arXiv preprint arXiv:2505.01457},
  year={2025}
}

@article{lopez2025enhancing,
  title={Enhancing Document VQA Models via Retrieval-Augmented Generation},
  author={L{\'o}pez, Eric and Llabr{\'e}s, Artemis and Valveny, Ernest},
  journal={arXiv preprint arXiv:2508.18984},
  year={2025}
}

@article{zhang2025patho,
  title={Patho-AgenticRAG: Towards Multimodal Agentic Retrieval-Augmented Generation for Pathology VLMs via Reinforcement Learning},
  author={Zhang, Wenchuan and Guo, Jingru and Zhang, Hengzhe and Zhang, Penghao and Chen, Jie and Zhang, Shuwan and Zhang, Zhang and Yi, Yuhao and Bu, Hong},
  journal={arXiv preprint arXiv:2508.02258},
  year={2025}
}

@article{cha2025generalized,
  title={Generalized Reinforcement Learning for Retriever-Specific Query Rewriter with Unstructured Real-World Documents},
  author={Cha, Sungguk and Kim, DongWook and Hahn, Taeseung and Kim, Mintae and Han, Youngsub and Jeon, Byoung-Ki},
  journal={arXiv preprint arXiv:2507.23242},
  year={2025}
}

@inproceedings{yu2025beyond,
  title={Beyond Relevance: Utility-Driven Retrieval for Visual Document Question Answering},
  author={Yu, Bihui and Wu, Gaowei and Yao, Zhuoya and Shi, Huiyang and Chen, Qi and Bu, Liping and Sun, Linzhuang and Wei, Jingxuan},
  booktitle={International Conference on Intelligent Computing},
  pages={382--393},
  year={2025},
  organization={Springer}
}

@article{xu2025mm,
  title={MM-R5: MultiModal Reasoning-Enhanced ReRanker via Reinforcement Learning for Document Retrieval},
  author={Xu, Mingjun and Dong, Jinhan and Hou, Jue and Wang, Zehui and Li, Sihang and Gao, Zhifeng and Zhong, Renxin and Cai, Hengxing},
  journal={arXiv preprint arXiv:2506.12364},
  year={2025}
}

@inproceedings{zhang2025bridging,
  title={Bridging Modalities: Improving Universal Multimodal Retrieval by Multimodal Large Language Models},
  author={Zhang, Xin and Zhang, Yanzhao and Xie, Wen and Li, Mingxin and Dai, Ziqi and Long, Dingkun and Xie, Pengjun and Zhang, Meishan and Li, Wenjie and Zhang, Min},
  booktitle={Proceedings of the Computer Vision and Pattern Recognition Conference},
  pages={9274--9285},
  year={2025}
}

@article{ma2025towards,
  title={Towards Storage-Efficient Visual Document Retrieval: An Empirical Study on Reducing Patch-Level Embeddings},
  author={Ma, Yubo and Li, Jinsong and Zang, Yuhang and Wu, Xiaobao and Dong, Xiaoyi and Zhang, Pan and Cao, Yuhang and Duan, Haodong and Wang, Jiaqi and Cao, Yixin and others},
  journal={arXiv preprint arXiv:2506.04997},
  year={2025}
}

@misc{xia2025db3,
      title     = {DB3 Team's Solution For Meta KDD Cup'25},
      author    = {Yikuan Xia and Jiazun Chen and Yirui Zhan and Suifeng Zhao and Weipeng Jiang and Chaorui Zhang and Wei Han and Bo Bai and Jun Gao},
      year      = {2025},
      eprint    = {2509.09681},
      archivePrefix = {arXiv},
      primaryClass = {cs.IR},
      doi       = {10.48550/arXiv.2509.09681},
      url       = {https://arxiv.org/abs/2509.09681}
}

@inproceedings{suri2025visdom,
  title={VisDoM: Multi-Document QA with Visually Rich Elements Using Multimodal Retrieval-Augmented Generation},
  author={Suri, Manan and Mathur, Puneet and Dernoncourt, Franck and Goswami, Kanika and Rossi, Ryan A and Manocha, Dinesh},
  booktitle={Proceedings of the 2025 Conference of the Nations of the Americas Chapter of the Association for Computational Linguistics: Human Language Technologies (Volume 1: Long Papers)},
  pages={6088--6109},
  year={2025}
}

@article{huang2025frag,
  title={FRAG: Frame Selection Augmented Generation for Long Video and Long Document Understanding},
  author={Huang, De-An and Radhakrishnan, Subhashree and Yu, Zhiding and Kautz, Jan},
  journal={arXiv preprint arXiv:2504.17447},
  year={2025}
}

@article{wang2025vidorag,
  title={ViDoRAG: Visual Document Retrieval-Augmented Generation via Dynamic Iterative Reasoning Agents},
  author={Wang, Qiuchen and Ding, Ruixue and Chen, Zehui and Wu, Weiqi and Wang, Shihang and Xie, Pengjun and Zhao, Feng},
  journal={arXiv preprint arXiv:2502.18017},
  year={2025}
}

@inproceedings{tanaka2025vdocrag,
  title={Vdocrag: Retrieval-augmented generation over visually-rich documents},
  author={Tanaka, Ryota and Iki, Taichi and Hasegawa, Taku and Nishida, Kyosuke and Saito, Kuniko and Suzuki, Jun},
  booktitle={Proceedings of the Computer Vision and Pattern Recognition Conference},
  pages={24827--24837},
  year={2025}
}

@article{lewis2020retrieval,
  title={Retrieval-augmented generation for knowledge-intensive nlp tasks},
  author={Lewis, Patrick and Perez, Ethan and Piktus, Aleksandra and Petroni, Fabio and Karpukhin, Vladimir and Goyal, Naman and K{\"u}ttler, Heinrich and Lewis, Mike and Yih, Wen-tau and Rockt{\"a}schel, Tim and others},
  journal={Advances in neural information processing systems},
  volume={33},
  pages={9459--9474},
  year={2020}
}

@inproceedings{jorensufficient,
  title={Sufficient Context: A New Lens on Retrieval Augmented Generation Systems},
  author={Joren, Hailey and Zhang, Jianyi and Ferng, Chun-Sung and Juan, Da-Cheng and Taly, Ankur and Rashtchian, Cyrus},
  booktitle={The Thirteenth International Conference on Learning Representations},
  yeal={2025}
}

@inproceedings{ye2024r2ag,
  title={R$^2$AG: Incorporating Retrieval Information into Retrieval Augmented Generation},
  author={Ye, Fuda and Li, Shuangyin and Zhang, Yongqi and Chen, Lei},
  booktitle={EMNLP (Findings)},
  year={2024}
}

@article{gupta2024comprehensive,
  title={A comprehensive survey of retrieval-augmented generation (rag): Evolution, current landscape and future directions},
  author={Gupta, Shailja and Ranjan, Rajesh and Singh, Surya Narayan},
  journal={arXiv preprint arXiv:2410.12837},
  year={2024}
}

@article{huang2024survey,
  title={A survey on retrieval-augmented text generation for large language models},
  author={Huang, Yizheng and Huang, Jimmy},
  journal={arXiv preprint arXiv:2404.10981},
  year={2024}
}

@article{cheng2025survey,
  title={A survey on knowledge-oriented retrieval-augmented generation},
  author={Cheng, Mingyue and Luo, Yucong and Ouyang, Jie and Liu, Qi and Liu, Huijie and Li, Li and Yu, Shuo and Zhang, Bohou and Cao, Jiawei and Ma, Jie and others},
  journal={arXiv preprint arXiv:2503.10677},
  year={2025}
}

@article{gondhalekar2025multifinrag,
  title={MultiFinRAG: An Optimized Multimodal Retrieval-Augmented Generation (RAG) Framework for Financial Question Answering},
  author={Gondhalekar, Chinmay and Patel, Urjitkumar and Yeh, Fang-Chun},
  journal={arXiv preprint arXiv:2506.20821},
  year={2025}
}

@article{zhao2025finragbench,
  title={FinRAGBench-V: A Benchmark for Multimodal RAG with Visual Citation in the Financial Domain},
  author={Zhao, Suifeng and Jin, Zhuoran and Li, Sujian and Gao, Jun},
  journal={arXiv preprint arXiv:2505.17471},
  year={2025}
}

@inproceedings{gokdemir2025hiperrag,
  title={HiPerRAG: High-Performance Retrieval Augmented Generation for Scientific Insights},
  author={Gokdemir, Ozan and Siebenschuh, Carlo and Brace, Alexander and Wells, Azton and Hsu, Brian and Hippe, Kyle and Setty, Priyanka and Ajith, Aswathy and Pauloski, J Gregory and Sastry, Varuni and others},
  booktitle={Proceedings of the Platform for Advanced Scientific Computing Conference},
  pages={1--13},
  year={2025}
}

@article{schneider2025collex,
  title={CollEX--A Multimodal Agentic RAG System Enabling Interactive Exploration of Scientific Collections},
  author={Schneider, Florian and Ahmadi, Narges Baba and Ahmadi, Niloufar Baba and Vogel, Iris and Semmann, Martin and Biemann, Chris},
  journal={arXiv preprint arXiv:2504.07643},
  year={2025}
}

@article{papageorgiou2025multimodal,
  title={A multimodal framework embedding retrieval-augmented generation with mllms for eurobarometer data},
  author={Papageorgiou, George and Sarlis, Vangelis and Maragoudakis, Manolis and Tjortjis, Christos},
  journal={AI},
  volume={6},
  number={3},
  pages={50},
  year={2025},
  publisher={MDPI}
}

@article{shereen2025one,
  title={One Pic is All it Takes: Poisoning Visual Document Retrieval Augmented Generation with a Single Image},
  author={Shereen, Ezzeldin and Ristea, Dan and McFadden, Shae and Hasircioglu, Burak and Mavroudis, Vasilios and Hicks, Chris},
  journal={arXiv preprint arXiv:2504.02132},
  year={2025}
}

@article{liu2025poisoned,
  title={Poisoned-MRAG: Knowledge Poisoning Attacks to Multimodal Retrieval Augmented Generation},
  author={Liu, Yinuo and Yuan, Zenghui and Tie, Guiyao and Shi, Jiawen and Zhou, Pan and Sun, Lichao and Gong, Neil Zhenqiang},
  journal={arXiv preprint arXiv:2503.06254},
  year={2025}
}

@article{gao2023retrieval,
  title={Retrieval-augmented generation for large language models: A survey},
  author={Gao, Yunfan and Xiong, Yun and Gao, Xinyu and Jia, Kangxiang and Pan, Jinliu and Bi, Yuxi and Dai, Yixin and Sun, Jiawei and Wang, Haofen and Wang, Haofen},
  journal={arXiv preprint arXiv:2312.10997},
  volume={2},
  number={1},
  year={2023}
}

@article{hu2024rag,
  title={Rag and rau: A survey on retrieval-augmented language model in natural language processing},
  author={Hu, Yucheng and Lu, Yuxing},
  journal={arXiv preprint arXiv:2404.19543},
  year={2024}
}

@article{zhao2024retrieval,
  title={Retrieval augmented generation (rag) and beyond: A comprehensive survey on how to make your llms use external data more wisely},
  author={Zhao, Siyun and Yang, Yuqing and Wang, Zilong and He, Zhiyuan and Qiu, Luna K and Qiu, Lili},
  journal={arXiv preprint arXiv:2409.14924},
  year={2024}
}

@article{church2024emerging,
  title={Emerging trends: a gentle introduction to RAG},
  author={Church, Kenneth Ward and Sun, Jiameng and Yue, Richard and Vickers, Peter and Saba, Walid and Chandrasekar, Raman},
  journal={Natural Language Engineering},
  volume={30},
  number={4},
  pages={870--881},
  year={2024},
  publisher={Cambridge University Press}
}

@article{zhao2023retrieving,
  title={Retrieving multimodal information for augmented generation: A survey},
  author={Zhao, Ruochen and Chen, Hailin and Wang, Weishi and Jiao, Fangkai and Do, Xuan Long and Qin, Chengwei and Ding, Bosheng and Guo, Xiaobao and Li, Minzhi and Li, Xingxuan and others},
  journal={arXiv preprint arXiv:2303.10868},
  year={2023}
}

@inproceedings{nandi2024visual,
  title={Visual Document Understanding: A Comparative Review of Modern Methods},
  author={Nandi, Kalyan and Sathya, S Siva},
  booktitle={International Conference on Computer Vision and Image Processing},
  pages={411--427},
  year={2024},
  organization={Springer}
}

@inproceedings{van2023document,
  title={Document understanding dataset and evaluation (dude)},
  author={Van Landeghem, Jordy and Tito, Rub{\`e}n and Borchmann, {\L}ukasz and Pietruszka, Micha{\l} and Joziak, Pawel and Powalski, Rafal and Jurkiewicz, Dawid and Coustaty, Micka{\"e}l and Anckaert, Bertrand and Valveny, Ernest and others},
  booktitle={Proceedings of the IEEE/CVF International Conference on Computer Vision},
  pages={19528--19540},
  year={2023}
}

@article{ding2025survey,
  title={A Survey on MLLM-based Visually Rich Document Understanding: Methods, Challenges, and Emerging Trends},
  author={Ding, Yihao and Luo, Siwen and Dai, Yue and Jiang, Yanbei and Li, Zechuan and Martin, Geoffrey and Peng, Yifan},
  journal={arXiv preprint arXiv:2507.09861},
  year={2025}
}

@article{li2024multimodal,
  title={Multimodal arxiv: A dataset for improving scientific comprehension of large vision-language models},
  author={Li, Lei and Wang, Yuqi and Xu, Runxin and Wang, Peiyi and Feng, Xiachong and Kong, Lingpeng and Liu, Qi},
  journal={arXiv preprint arXiv:2403.00231},
  year={2024}
}

@inproceedings{mathew2021docvqa,
  title={Docvqa: A dataset for vqa on document images},
  author={Mathew, Minesh and Karatzas, Dimosthenis and Jawahar, CV},
  booktitle={Proceedings of the IEEE/CVF winter conference on applications of computer vision},
  pages={2200--2209},
  year={2021}
}

@inproceedings{mathew2022infographicvqa,
  title={Infographicvqa},
  author={Mathew, Minesh and Bagal, Viraj and Tito, Rub{\`e}n and Karatzas, Dimosthenis and Valveny, Ernest and Jawahar, CV},
  booktitle={Proceedings of the IEEE/CVF Winter Conference on Applications of Computer Vision},
  pages={1697--1706},
  year={2022}
}

@article{zhu2021tat,
  title={TAT-QA: A question answering benchmark on a hybrid of tabular and textual content in finance},
  author={Zhu, Fengbin and Lei, Wenqiang and Huang, Youcheng and Wang, Chao and Zhang, Shuo and Lv, Jiancheng and Feng, Fuli and Chua, Tat-Seng},
  journal={arXiv preprint arXiv:2105.07624},
  year={2021}
}

@article{cho2024typos,
  title={Typos that Broke the RAG's Back: Genetic Attack on RAG Pipeline by Simulating Documents in the Wild via Low-level Perturbations},
  author={Cho, Sukmin and Jeong, Soyeong and Seo, Jeongyeon and Hwang, Taeho and Park, Jong C},
  journal={arXiv preprint arXiv:2404.13948},
  year={2024}
}

@inproceedings{nazary2025poison,
  title={Poison-rag: Adversarial data poisoning attacks on retrieval-augmented generation in recommender systems},
  author={Nazary, Fatemeh and Deldjoo, Yashar and Noia, Tommaso di},
  booktitle={European Conference on Information Retrieval},
  pages={239--251},
  year={2025},
  organization={Springer}
}

@article{jiang2024rag,
  title={Rag-thief: Scalable extraction of private data from retrieval-augmented generation applications with agent-based attacks},
  author={Jiang, Changyue and Pan, Xudong and Hong, Geng and Bao, Chenfu and Yang, Min},
  journal={arXiv preprint arXiv:2411.14110},
  year={2024}
}

@article{xian2024understanding,
  title={Understanding Data Poisoning Attacks for RAG: Insights and Algorithms},
  author={Xian, Xun and Wang, Tong and You, Liwen and Qi, Yanjun},
  year={2024}
}

@inproceedings{zhu2022towards,
  title={Towards complex document understanding by discrete reasoning},
  author={Zhu, Fengbin and Lei, Wenqiang and Feng, Fuli and Wang, Chao and Zhang, Haozhou and Chua, Tat-Seng},
  booktitle={Proceedings of the 30th ACM International Conference on Multimedia},
  pages={4857--4866},
  year={2022}
}

@article{d2020fquad,
  title={FQuAD: French question answering dataset},
  author={d'Hoffschmidt, Martin and Belblidia, Wacim and Brendl{\'e}, Tom and Heinrich, Quentin and Vidal, Maxime},
  journal={arXiv preprint arXiv:2002.06071},
  year={2020}
}

@inproceedings{tanaka2021visualmrc,
  title={Visualmrc: Machine reading comprehension on document images},
  author={Tanaka, Ryota and Nishida, Kyosuke and Yoshida, Sen},
  booktitle={Proceedings of the AAAI Conference on Artificial Intelligence},
  volume={35},
  number={15},
  pages={13878--13888},
  year={2021}
}

@article{masry2022chartqa,
  title={Chartqa: A benchmark for question answering about charts with visual and logical reasoning},
  author={Masry, Ahmed and Long, Do Xuan and Tan, Jia Qing and Joty, Shafiq and Hoque, Enamul},
  journal={arXiv preprint arXiv:2203.10244},
  year={2022}
}

@inproceedings{methani2020plotqa,
  title={Plotqa: Reasoning over scientific plots},
  author={Methani, Nitesh and Ganguly, Pritha and Khapra, Mitesh M and Kumar, Pratyush},
  booktitle={Proceedings of the ieee/cvf winter conference on applications of computer vision},
  pages={1527--1536},
  year={2020}
}

@inproceedings{tanaka2023slidevqa,
  title={Slidevqa: A dataset for document visual question answering on multiple images},
  author={Tanaka, Ryota and Nishida, Kyosuke and Nishida, Kosuke and Hasegawa, Taku and Saito, Itsumi and Saito, Kuniko},
  booktitle={Proceedings of the AAAI Conference on Artificial Intelligence},
  volume={37},
  number={11},
  pages={13636--13645},
  year={2023}
}

@article{hui2024uda,
  title={Uda: A benchmark suite for retrieval augmented generation in real-world document analysis},
  author={Hui, Yulong and Lu, Yao and Zhang, Huanchen},
  journal={Advances in Neural Information Processing Systems},
  volume={37},
  pages={67200--67217},
  year={2024}
}

@article{pramanick2024spiqa,
  title={Spiqa: A dataset for multimodal question answering on scientific papers},
  author={Pramanick, Shraman and Chellappa, Rama and Venugopalan, Subhashini},
  journal={Advances in Neural Information Processing Systems},
  volume={37},
  pages={118807--118833},
  year={2024}
}

@article{saikh2022scienceqa,
  title={Scienceqa: A novel resource for question answering on scholarly articles},
  author={Saikh, Tanik and Ghosal, Tirthankar and Mittal, Amish and Ekbal, Asif and Bhattacharyya, Pushpak},
  journal={International Journal on Digital Libraries},
  volume={23},
  number={3},
  pages={289--301},
  year={2022},
  publisher={Springer}
}

@article{li2024benchmarking,
  title={Benchmarking multimodal retrieval augmented generation with dynamic vqa dataset and self-adaptive planning agent},
  author={Li, Yangning and Li, Yinghui and Wang, Xinyu and Jiang, Yong and Zhang, Zhen and Zheng, Xinran and Wang, Hui and Zheng, Hai-Tao and Yu, Philip S and Huang, Fei and others},
  journal={arXiv preprint arXiv:2411.02937},
  year={2024}
}

@article{christen2023review,
  title={A review of the F-measure: its history, properties, criticism, and alternatives},
  author={Christen, Peter and Hand, David J and Kirielle, Nishadi},
  journal={ACM Computing Surveys},
  volume={56},
  number={3},
  pages={1--24},
  year={2023},
  publisher={ACM New York, NY}
}

@inproceedings{omar-etal-2024-multi,
    title = "Multi-Level Information Retrieval Augmented Generation for Knowledge-based Visual Question Answering",
    author = "Adjali, Omar  and
      Ferret, Olivier  and
      Ghannay, Sahar  and
      Le Borgne, Herv{\'e}",
    url = "https://aclanthology.org/2024.emnlp-main.922/",
}

@misc{nguyen2024,
      title={Multimodal Learned Sparse Retrieval with Probabilistic Expansion Control}, 
      author={Thong Nguyen and Mariya Hendriksen and Andrew Yates and Maarten de Rijke},
      year={2024},
      eprint={2402.17535},
      archivePrefix={arXiv},
      primaryClass={cs.IR},
      url={https://arxiv.org/abs/2402.17535}, 
}

@inproceedings{rajpurkar-etal-2016-squad,
    title = "{SQ}u{AD}: 100,000+ Questions for Machine Comprehension of Text",
    author = "Rajpurkar, Pranav  and
      Zhang, Jian  and
      Lopyrev, Konstantin  and
      Liang, Percy",
    editor = "Su, Jian  and
      Duh, Kevin  and
      Carreras, Xavier",
    booktitle = "Proceedings of the 2016 Conference on Empirical Methods in Natural Language Processing",
    month = nov,
    year = "2016",
    address = "Austin, Texas",
    publisher = "Association for Computational Linguistics",
    url = "https://aclanthology.org/D16-1264/",
    doi = "10.18653/v1/D16-1264",
    pages = "2383--2392"
}

@inproceedings{papineni2002bleu,
  title={Bleu: a method for automatic evaluation of machine translation},
  author={Papineni, Kishore and Roukos, Salim and Ward, Todd and Zhu, Wei-Jing},
  booktitle={Proceedings of the 40th annual meeting of the Association for Computational Linguistics},
  pages={311--318},
  year={2002}
}

@inproceedings{lin2004rouge,
  title={ROUGE: A Package for Automatic Evaluation of Summaries},
  author={Lin, Chin-Yew},
  booktitle={Text Summarization Branches Out: Proceedings of the ACL Workshop},
  pages={74--81},
  year={2004}
}

@inproceedings{banerjee2005meteor,
  title={METEOR: An Automatic Metric for MT Evaluation with Improved Correlation with Human Judgments},
  author={Banerjee, Satanjeev and Lavie, Alon},
  booktitle={Proceedings of the ACL Workshop on Intrinsic and Extrinsic Evaluation Measures},
  pages={65--72},
  year={2005}
}

@inproceedings{zhang2024exploring,
  title={Exploring the capabilities of large multimodal models on dense text},
  author={Zhang, Shuo and Yang, Biao and Li, Zhang and Ma, Zhiyin and Liu, Yuliang and Bai, Xiang},
  booktitle={International Conference on Document Analysis and Recognition},
  pages={281--298},
  year={2024},
  organization={Springer}
}

@article{chen2024mmr,
  title={Mmr: Evaluating reading ability of large multimodal models},
  author={Chen, Jian and Zhang, Ruiyi and Zhou, Yufan and Rossi, Ryan and Gu, Jiuxiang and Chen, Changyou},
  journal={arXiv preprint arXiv:2408.14594},
  year={2024}
}

@inproceedings{devlin2019bert,
  title     = {BERT: Pre-training of Deep Bidirectional Transformers for Language Understanding},
  author    = {Devlin, Jacob and Chang, Ming-Wei and Lee, Kenton and Toutanova, Kristina},
  booktitle = {Proceedings of the 2019 Conference of the North {A}merican Chapter of the Association for Computational Linguistics: Human Language Technologies (NAACL-HLT)},
  year      = {2019},
  pages     = {4171--4186},
  publisher = {Association for Computational Linguistics},
  url       = {https://aclanthology.org/N19-1423}
}

@article{liu2019roberta,
  title   = {RoBERTa: A Robustly Optimized BERT Pretraining Approach},
  author  = {Liu, Yinhan and Ott, Myle and Goyal, Naman and Du, Jingfei and Joshi, Mandar and Chen, Danqi and Levy, Omer and Lewis, Mike and Zettlemoyer, Luke and Stoyanov, Veselin},
  journal = {arXiv preprint arXiv:1907.11692},
  year    = {2019},
  url     = {https://arxiv.org/abs/1907.11692}
}

@article{nguyen2025serval,
  title={SERVAL: Surprisingly Effective Zero-Shot Visual Document Retrieval Powered by Large Vision and Language Models},
  author={Nguyen, Thong and Lei, Yibin and Ju, Jia-Huei and Yates, Andrew},
  journal={arXiv preprint arXiv:2509.15432},
  year={2025}
}

@article{xiao2025metaembed,
  title={MetaEmbed: Scaling Multimodal Retrieval at Test-Time with Flexible Late Interaction},
  author={Xiao, Zilin and Ma, Qi and Gu, Mengting and Chen, Chun-cheng Jason and Chen, Xintao and Ordonez, Vicente and Mohan, Vijai},
  journal={arXiv preprint arXiv:2509.18095},
  year={2025}
}

@article{yan2025docpruner,
  title={DocPruner: A Storage-Efficient Framework for Multi-Vector Visual Document Retrieval via Adaptive Patch-Level Embedding Pruning},
  author={Yan, Yibo and Xu, Guangwei and Zou, Xin and Liu, Shuliang and Kwok, James and Hu, Xuming},
  journal={arXiv preprint arXiv:2509.23883},
  year={2025}
}

@inproceedings{biten2019scene,
  title={Scene text visual question answering},
  author={Biten, Ali Furkan and Tito, Ruben and Mafla, Andres and Gomez, Lluis and Rusinol, Mar{\c{c}}al and Valveny, Ernest and Jawahar, CV and Karatzas, Dimosthenis},
  booktitle={Proceedings of the IEEE/CVF international conference on computer vision},
  pages={4291--4301},
  year={2019}
}

@article{sellers1980theory,
  title={The theory and computation of evolutionary distances: pattern recognition},
  author={Sellers, Peter H},
  journal={Journal of algorithms},
  volume={1},
  number={4},
  pages={359--373},
  year={1980},
  publisher={Elsevier}
}

@inproceedings{cormack2009reciprocal,
  title={Reciprocal rank fusion outperforms condorcet and individual rank learning methods},
  author={Cormack, Gordon V and Clarke, Charles LA and Buettcher, Stefan},
  booktitle={Proceedings of the 32nd international ACM SIGIR conference on Research and development in information retrieval},
  pages={758--759},
  year={2009}
}

@inproceedings{calumby2017rank,
  title={Rank Fusion and Multimodal Per-topic Adaptiveness for Diverse Image Retrieval.},
  author={Calumby, Rodrigo Tripodi and do Carmo Araujo, Iago Breno Alves and Cordeiro, Felipe Souza and Bertoni, Fabiana and Canuto, S{\'e}rgio D and Bel{\'e}m, Fabiano and Gon{\c{c}}alves, Marcos Andr{\'e} and Dourado, {\'I}caro C and Munoz, Javier AV and Li, Lin and others},
  booktitle={MediaEval},
  year={2017}
}

@article{jarvelin2002cumulated,
  title={Cumulated gain-based evaluation of IR techniques},
  author={J{\"a}rvelin, Kalervo and Kek{\"a}l{\"a}inen, Jaana},
  journal={ACM Transactions on Information Systems (TOIS)},
  volume={20},
  number={4},
  pages={422--446},
  year={2002},
  publisher={ACM New York, NY, USA}
}

@inproceedings{lcvenshtcin1966binary,
  title={Binary coors capable or ‘correcting deletions, insertions, and reversals},
  author={Lcvenshtcin, VI},
  booktitle={Soviet physics-doklady},
  volume={10},
  number={8},
  year={1966}
}

@article{khosla2020supervised,
  title={Supervised contrastive learning},
  author={Khosla, Prannay and Teterwak, Piotr and Wang, Chen and Sarna, Aaron and Tian, Yonglong and Isola, Phillip and Maschinot, Aaron and Liu, Ce and Krishnan, Dilip},
  journal={Advances in neural information processing systems},
  volume={33},
  pages={18661--18673},
  year={2020}
}

@inproceedings{wang2021understanding,
  title={Understanding the behaviour of contrastive loss},
  author={Wang, Feng and Liu, Huaping},
  booktitle={Proceedings of the IEEE/CVF conference on computer vision and pattern recognition},
  pages={2495--2504},
  year={2021}
}

@article{wangrecon,
  title={RECON: Multimodal GraphRAG for Visually Rich Documents with Intra-Page Reflection and Inter-Page Connection},
  author={Wang, Yi-Cheng and Chen, Chu-Song},
  year={2025}
}

@article{sourati2025lad,
  title={LAD-RAG: Layout-aware Dynamic RAG for Visually-Rich Document Understanding},
  author={Sourati, Zhivar and Wang, Zheng and Liu, Marianne Menglin and Hu, Yazhe and Guo, Mengqing and Bharadwaj, Sujeeth and Han, Kyu and Sheng, Tao and Ravi, Sujith and Dehghani, Morteza and others},
  journal={arXiv preprint arXiv:2510.07233},
  year={2025}
}

@article{kim2025hybrid,
  title={Hybrid-Vector Retrieval for Visually Rich Documents: Combining Single-Vector Efficiency and Multi-Vector Accuracy},
  author={Kim, Juyeon and Lee, Geon and Choi, Dongwon and Kim, Taeuk and Shin, Kijung},
  journal={arXiv preprint arXiv:2510.22215},
  year={2025}
}

@inproceedings{zhang2025dream,
  title={DREAM: Integrating Hierarchical Multimodal Retrieval with Multi-page Multimodal Language Model for Documents VQA},
  author={Zhang, Jinxu and Fan, Qiyuan and Yu, Yongqi and Zhang, Yu},
  booktitle={Proceedings of the 33rd ACM International Conference on Multimedia},
  pages={4213--4221},
  year={2025}
}

@inproceedings{wu2025mara,
  title={MARA: A Multimodal Adaptive Retrieval-Augmented Framework for Document Question Answering},
  author={Wu, Hui and Zhai, Haoquan and Li, Yuchen and Cai, Hengyi and Zhang, Peirong and Zhang, Yidan and Wang, Lei and Wang, Chunle and Hou, Yingyan and Wang, Shuaiqiang and others},
  booktitle={Proceedings of the 33rd ACM International Conference on Multimedia},
  pages={4329--4338},
  year={2025}
}

@inproceedings{chen2025hear,
  title={HEAR: A Holistic Extraction and Agentic Reasoning Framework for Document Understanding},
  author={Chen, Longfeng and Xiao, Zheng and Wang, Juyuan and Huang, Zeyu and Zeng, Yawen and Xu, Jin},
  booktitle={Proceedings of the 33rd ACM International Conference on Multimedia},
  pages={14376--14382},
  year={2025}
}

@article{bach2025hierarchical,
  title={Hierarchical Patch Compression for ColPali: Efficient Multi-Vector Document Retrieval with Dynamic Pruning and Quantization},
  author={Bach, Duong},
  journal={arXiv preprint arXiv:2506.21601},
  year={2025}
}

@article{li2025regionrag,
  title={RegionRAG: Region-level Retrieval-Augumented Generation for Visually-Rich Documents},
  author={Li, Yinglu and Lu, Zhiying and Liu, Zhihang and Liu, Chuanbin and Xie, Hongtao},
  journal={arXiv preprint arXiv:2510.27261},
  year={2025}
}

@inproceedings{lim2025distilling,
  title={Distilling Cross-Modal Knowledge into Domain-Specific Retrievers for Enhanced Industrial Document Understanding},
  author={Lim, Jinhyeong and Shin, Jeongwan and Lee, Seeun and Kim, Seongdeok and Choi, Joungsu and Kim, Jongbae and Jung, Chun Hwan and Kang, Youjin},
  booktitle={Proceedings of the 2025 Conference on Empirical Methods in Natural Language Processing: Industry Track},
  pages={2551--2563},
  year={2025}
}

@inproceedings{masry2025colmate,
  title={COLMATE: Contrastive Late Interaction and Masked Text for Multimodal Document Retrieval},
  author={Masry, Ahmed and Thakkar, Megh and Bechard, Patrice and Madhusudhan, Sathwik Tejaswi and Awal, Rabiul and Mishra, Shambhavi and Suresh, Akshay Kalkunte and Daruru, Srivatsava and Hoque, Enamul and Gella, Spandana and others},
  booktitle={Proceedings of the 2025 Conference on Empirical Methods in Natural Language Processing: Industry Track},
  pages={2071--2080},
  year={2025}
}

@inproceedings{yun2025lilac,
  title={LILaC: Late Interacting in Layered Component Graph for Open-domain Multimodal Multihop Retrieval},
  author={Yun, Joohyung and Lee, Doyup and Han, Wook-Shin},
  booktitle={Proceedings of the 2025 Conference on Empirical Methods in Natural Language Processing},
  pages={20551--20570},
  year={2025}
}

@article{tong2025hkrag,
  title={HKRAG: Holistic Knowledge Retrieval-Augmented Generation Over Visually-Rich Documents},
  author={Tong, Anyang and Niu, Xiang and Liu, ZhiPing and Tian, Chang and Wei, Yanyan and Shi, Zenglin and Wang, Meng},
  journal={arXiv preprint arXiv:2511.20227},
  year={2025}
}

@article{liu2025resolving,
  title={Resolving Evidence Sparsity: Agentic Context Engineering for Long-Document Understanding},
  author={Liu, Keliang and Chen, Zizhi and Li, Mingcheng and Tang, Jingqun and Yang, Dingkang and Zhang, Lihua},
  journal={arXiv preprint arXiv:2511.22850},
  year={2025}
}

@article{georgiou2025spatially,
  title={Spatially-Grounded Document Retrieval via Patch-to-Region Relevance Propagation},
  author={Georgiou, Agathoklis},
  journal={arXiv preprint arXiv:2512.02660},
  year={2025}
}

@inproceedings{maleki2024ai,
  title={AI hallucinations: a misnomer worth clarifying},
  author={Maleki, Negar and Padmanabhan, Balaji and Dutta, Kaushik},
  booktitle={2024 IEEE conference on artificial intelligence (CAI)},
  pages={133--138},
  year={2024},
  organization={IEEE}
}

@article{liu2024survey,
  title={A survey on hallucination in large vision-language models},
  author={Liu, Hanchao and Xue, Wenyuan and Chen, Yifei and Chen, Dapeng and Zhao, Xiutian and Wang, Ke and Hou, Liping and Li, Rongjun and Peng, Wei},
  journal={arXiv preprint arXiv:2402.00253},
  year={2024}
}

@article{wang2024mitigating,
  title={Mitigating hallucinations in large vision-language models with instruction contrastive decoding},
  author={Wang, Xintong and Pan, Jingheng and Ding, Liang and Biemann, Chris},
  journal={arXiv preprint arXiv:2403.18715},
  year={2024}
}

@article{xu2024benchmarking,
  title={Benchmarking benchmark leakage in large language models},
  author={Xu, Ruijie and Wang, Zengzhi and Fan, Run-Ze and Liu, Pengfei},
  journal={arXiv preprint arXiv:2404.18824},
  year={2024}
}

@inproceedings{hu2025vlsbench,
  title={Vlsbench: Unveiling visual leakage in multimodal safety},
  author={Hu, Xuhao and Liu, Dongrui and Li, Hao and Huang, Xuan-Jing and Shao, Jing},
  booktitle={Proceedings of the 63rd Annual Meeting of the Association for Computational Linguistics (Volume 1: Long Papers)},
  pages={8285--8316},
  year={2025}
}

@article{zhou2025lessleak,
  title={Lessleak-bench: A first investigation of data leakage in llms across 83 software engineering benchmarks},
  author={Zhou, Xin and Weyssow, Martin and Widyasari, Ratnadira and Zhang, Ting and He, Junda and Lyu, Yunbo and Chang, Jianming and Zhang, Beiqi and Huang, Dan and Lo, David},
  journal={arXiv preprint arXiv:2502.06215},
  year={2025}
}

@article{xu2024benchmark,
  title={Benchmark data contamination of large language models: A survey},
  author={Xu, Cheng and Guan, Shuhao and Greene, Derek and Kechadi, M and others},
  journal={arXiv preprint arXiv:2406.04244},
  year={2024}
}

@inproceedings{deng2024investigating,
  title={Investigating data contamination in modern benchmarks for large language models},
  author={Deng, Chunyuan and Zhao, Yilun and Tang, Xiangru and Gerstein, Mark and Cohan, Arman},
  booktitle={Proceedings of the 2024 Conference of the North American Chapter of the Association for Computational Linguistics: Human Language Technologies (Volume 1: Long Papers)},
  pages={8706--8719},
  year={2024}
}

@article{riedler2024beyond,
  title={Beyond text: Optimizing rag with multimodal inputs for industrial applications},
  author={Riedler, Monica and Langer, Stefan},
  journal={arXiv preprint arXiv:2410.21943},
  year={2024}
}

@inproceedings{liu2024optimizing,
  title={Optimizing rag techniques for automotive industry pdf chatbots: A case study with locally deployed ollama modelsoptimizing rag techniques based on locally deployed ollama modelsa case study with locally deployed ollama models},
  author={Liu, Fei and Kang, Zejun and Han, Xing},
  booktitle={Proceedings of the 2024 3rd International Conference on Artificial Intelligence and Intelligent Information Processing},
  pages={152--159},
  year={2024}
}

@article{bourdin2025agile,
  title={An Agile Method for Implementing Retrieval Augmented Generation Tools in Industrial SMEs},
  author={Bourdin, Mathieu and Neumann, Anas and Paviot, Thomas and Pellerin, Robert and Lamouri, Samir},
  journal={arXiv preprint arXiv:2508.21024},
  year={2025}
}

@article{brehme2025retrieval,
  title={Retrieval-Augmented Generation in Industry: An Interview Study on Use Cases, Requirements, Challenges, and Evaluation},
  author={Brehme, Lorenz and Dornauer, Benedikt and Str{\"o}hle, Thomas and Ehrhart, Maximilian and Breu, Ruth},
  journal={arXiv preprint arXiv:2508.14066},
  year={2025}
}

@article{chen2025application,
  title={Application of retrieval-augmented generation for interactive industrial knowledge management via a large language model},
  author={Chen, Lun-Chi and Pardeshi, Mayuresh Sunil and Liao, Yi-Xiang and Pai, Kai-Chih},
  journal={Computer Standards \& Interfaces},
  volume={94},
  pages={103995},
  year={2025},
  publisher={Elsevier}
}

@article{shi2025enhancing,
  title={Enhancing retrieval-augmented generation for interoperable industrial knowledge representation and inference toward cognitive digital twins},
  author={Shi, Dachuan and Li, Jianzhang and Meyer, Olga and Bauernhansl, Thomas},
  journal={Computers in Industry},
  volume={171},
  pages={104330},
  year={2025},
  publisher={Elsevier}
}

@article{an2024golden,
  title={Golden-retriever: high-fidelity agentic retrieval augmented generation for industrial knowledge base},
  author={An, Zhiyu and Ding, Xianzhong and Fu, Yen-Chun and Chu, Cheng-Chung and Li, Yan and Du, Wan},
  journal={arXiv preprint arXiv:2408.00798},
  year={2024}
}

@misc{RAG-Anything,
  title = {RAG-Anything},
  author = {Intelligence Lab@HKU Data},
  year = {2025},
  url = {https://github.com/HKUDS/RAG-Anything},
  note = {Accessed 2025-12-26}
}

@misc{ragflow,
  title = {ragflow},
  author = {InfiniFlow},
  year = {2023},
  url = {https://github.com/infiniflow/ragflow},
  note = {Accessed 2025-12-26}
}

@misc{AutoRAG,
  title = {AutoRAG},
  author = {Markr AI},
  year = {2024},
  url = {https://github.com/Marker-Inc-Korea/AutoRAG},
  note = {Accessed 2025-12-26}
}

@misc{LightRAG,
  title = {LightRAG},
  author = {Intelligence Lab@HKU Data},
  year = {2024},
  url = {https://github.com/HKUDS/LightRAG},
  note = {Accessed 2025-12-26}
}

@misc{raglite,
  title = {raglite},
  author = {Superlinear},
  year = {2024},
  url = {https://github.com/superlinear-ai/raglite},
  note = {Accessed 2025-12-26}
}

@misc{LlamaIndex,
  title = {LlamaIndex},
  author = {LlamaIndex},
  year = {2025},
  doi = {},
  url = {https://github.com/jerryjliu/llama_index}
}

@article{bruckhaus2024rag,
  title={Rag does not work for enterprises},
  author={Bruckhaus, Tilmann},
  journal={arXiv preprint arXiv:2406.04369},
  year={2024}
}

@article{muller2025data,
  title={Data Quality Challenges in Retrieval-Augmented Generation},
  author={M{\"u}ller, Leopold and Holstein, Joshua and Bause, Sarah and Satzger, Gerhard and K{\"u}hl, Niklas},
  journal={arXiv preprint arXiv:2510.00552},
  year={2025}
}

@inproceedings{liu2024grounding,
  title={Grounding dino: Marrying dino with grounded pre-training for open-set object detection},
  author={Liu, Shilong and Zeng, Zhaoyang and Ren, Tianhe and Li, Feng and Zhang, Hao and Yang, Jie and Jiang, Qing and Li, Chunyuan and Yang, Jianwei and Su, Hang and others},
  booktitle={European conference on computer vision},
  pages={38--55},
  year={2024},
  organization={Springer}
}

@inproceedings{deng2021transvg,
  title={Transvg: End-to-end visual grounding with transformers},
  author={Deng, Jiajun and Yang, Zhengyuan and Chen, Tianlang and Zhou, Wengang and Li, Houqiang},
  booktitle={Proceedings of the IEEE/CVF international conference on computer vision},
  pages={1769--1779},
  year={2021}
}

@article{xiao2024towards,
  title={Towards visual grounding: A survey},
  author={Xiao, Linhui and Yang, Xiaoshan and Lan, Xiangyuan and Wang, Yaowei and Xu, Changsheng},
  journal={arXiv preprint arXiv:2412.20206},
  year={2024}
}

@article{peng2024graph,
  title={Graph retrieval-augmented generation: A survey},
  author={Peng, Boci and Zhu, Yun and Liu, Yongchao and Bo, Xiaohe and Shi, Haizhou and Hong, Chuntao and Zhang, Yan and Tang, Siliang},
  journal={arXiv preprint arXiv:2408.08921},
  year={2024}
}

@inproceedings{procko2024graph,
  title={Graph retrieval-augmented generation for large language models: A survey},
  author={Procko, Tyler Thomas and Ochoa, Omar},
  booktitle={2024 Conference on AI, Science, Engineering, and Technology (AIxSET)},
  pages={166--169},
  year={2024},
  organization={IEEE}
}

@article{zhang2025survey,
  title={A survey of graph retrieval-augmented generation for customized large language models},
  author={Zhang, Qinggang and Chen, Shengyuan and Bei, Yuanchen and Yuan, Zheng and Zhou, Huachi and Hong, Zijin and Chen, Hao and Xiao, Yilin and Zhou, Chuang and Dong, Junnan and others},
  journal={arXiv preprint arXiv:2501.13958},
  year={2025}
}

@article{li2024simple,
  title={Simple is effective: The roles of graphs and large language models in knowledge-graph-based retrieval-augmented generation},
  author={Li, Mufei and Miao, Siqi and Li, Pan},
  journal={arXiv preprint arXiv:2410.20724},
  year={2024}
}

@article{he2024g,
  title={G-retriever: Retrieval-augmented generation for textual graph understanding and question answering},
  author={He, Xiaoxin and Tian, Yijun and Sun, Yifei and Chawla, Nitesh and Laurent, Thomas and LeCun, Yann and Bresson, Xavier and Hooi, Bryan},
  journal={Advances in Neural Information Processing Systems},
  volume={37},
  pages={132876--132907},
  year={2024}
}

@article{ma2024think,
  title={Think-on-graph 2.0: Deep and faithful large language model reasoning with knowledge-guided retrieval augmented generation},
  author={Ma, Shengjie and Xu, Chengjin and Jiang, Xuhui and Li, Muzhi and Qu, Huaren and Yang, Cehao and Mao, Jiaxin and Guo, Jian},
  journal={arXiv preprint arXiv:2407.10805},
  year={2024}
}

@article{edge2024local,
  title={From local to global: A graph rag approach to query-focused summarization},
  author={Edge, Darren and Trinh, Ha and Cheng, Newman and Bradley, Joshua and Chao, Alex and Mody, Apurva and Truitt, Steven and Metropolitansky, Dasha and Ness, Robert Osazuwa and Larson, Jonathan},
  journal={arXiv preprint arXiv:2404.16130},
  year={2024}
}

@article{mavromatis2024gnn,
  title={Gnn-rag: Graph neural retrieval for large language model reasoning},
  author={Mavromatis, Costas and Karypis, George},
  journal={arXiv preprint arXiv:2405.20139},
  year={2024}
}

@article{luo2025gfm,
  title={GFM-RAG: graph foundation model for retrieval augmented generation},
  author={Luo, Linhao and Zhao, Zicheng and Haffari, Gholamreza and Phung, Dinh and Gong, Chen and Pan, Shirui},
  journal={arXiv preprint arXiv:2502.01113},
  year={2025}
}

@article{huang2025ket,
  title={Ket-rag: A cost-efficient multi-granular indexing framework for graph-rag},
  author={Huang, Yiqian and Zhang, Shiqi and Xiao, Xiaokui},
  journal={arXiv preprint arXiv:2502.09304},
  year={2025}
}

@article{guo2024lightrag,
  title={Lightrag: Simple and fast retrieval-augmented generation},
  author={Guo, Zirui and Xia, Lianghao and Yu, Yanhua and Ao, Tu and Huang, Chao},
  journal={arXiv preprint arXiv:2410.05779},
  year={2024}
}

@article{gutierrez2025rag,
  title={From rag to memory: Non-parametric continual learning for large language models},
  author={Guti{\'e}rrez, Bernal Jim{\'e}nez and Shu, Yiheng and Qi, Weijian and Zhou, Sizhe and Su, Yu},
  journal={arXiv preprint arXiv:2502.14802},
  year={2025}
}

@article{singh2025agentic,
  title={Agentic retrieval-augmented generation: A survey on agentic rag},
  author={Singh, Aditi and Ehtesham, Abul and Kumar, Saket and Khoei, Tala Talaei},
  journal={arXiv preprint arXiv:2501.09136},
  year={2025}
}

@article{li2025survey,
  title={A survey of personalization: From rag to agent},
  author={Li, Xiaopeng and Jia, Pengyue and Xu, Derong and Wen, Yi and Zhang, Yingyi and Zhang, Wenlin and Wang, Wanyu and Wang, Yichao and Du, Zhaocheng and Li, Xiangyang and others},
  journal={arXiv preprint arXiv:2504.10147},
  year={2025}
}

@inproceedings{chang2025main,
  title={Main-rag: Multi-agent filtering retrieval-augmented generation},
  author={Chang, Chia-Yuan and Jiang, Zhimeng and Rakesh, Vineeth and Pan, Menghai and Yeh, Chin-Chia Michael and Wang, Guanchu and Hu, Mingzhi and Xu, Zhichao and Zheng, Yan and Das, Mahashweta and others},
  booktitle={Proceedings of the 63rd Annual Meeting of the Association for Computational Linguistics (Volume 1: Long Papers)},
  pages={2607--2622},
  year={2025}
}

@article{nguyen2025ma,
  title={MA-RAG: Multi-Agent Retrieval-Augmented Generation via Collaborative Chain-of-Thought Reasoning},
  author={Nguyen, Thang and Chin, Peter and Tai, Yu-Wing},
  journal={arXiv preprint arXiv:2505.20096},
  year={2025}
}

@article{chen2025improving,
  title={Improving retrieval-augmented generation through multi-agent reinforcement learning},
  author={Chen, Yiqun and Yan, Lingyong and Sun, Weiwei and Ma, Xinyu and Zhang, Yi and Wang, Shuaiqiang and Yin, Dawei and Yang, Yiming and Mao, Jiaxin},
  journal={arXiv preprint arXiv:2501.15228},
  year={2025}
}

@inproceedings{jang2024rag,
  title={Au-rag: Agent-based universal retrieval augmented generation},
  author={Jang, Jisoo and Li, Wen-Syan},
  booktitle={Proceedings of the 2024 Annual International ACM SIGIR Conference on Research and Development in Information Retrieval in the Asia Pacific Region},
  pages={2--11},
  year={2024}
}

@article{yang2025qwen3,
  title={Qwen3 technical report},
  author={Yang, An and Li, Anfeng and Yang, Baosong and Zhang, Beichen and Hui, Binyuan and Zheng, Bo and Yu, Bowen and Gao, Chang and Huang, Chengen and Lv, Chenxu and others},
  journal={arXiv preprint arXiv:2505.09388},
  year={2025}
}

@article{achiam2023gpt,
  title={Gpt-4 technical report},
  author={Achiam, Josh and Adler, Steven and Agarwal, Sandhini and Ahmad, Lama and Akkaya, Ilge and Aleman, Florencia Leoni and Almeida, Diogo and Altenschmidt, Janko and Altman, Sam and Anadkat, Shyamal and others},
  journal={arXiv preprint arXiv:2303.08774},
  year={2023}
}

@article{team2023gemini,
  title={Gemini: a family of highly capable multimodal models},
  author={Team, Gemini and Anil, Rohan and Borgeaud, Sebastian and Alayrac, Jean-Baptiste and Yu, Jiahui and Soricut, Radu and Schalkwyk, Johan and Dai, Andrew M and Hauth, Anja and Millican, Katie and others},
  journal={arXiv preprint arXiv:2312.11805},
  year={2023}
}

@article{peng2025unidoc,
  title={UNIDOC-BENCH: A Unified Benchmark for Document-Centric Multimodal RAG},
  author={Peng, Xiangyu and Qin, Can and Chen, Zeyuan and Xu, Ran and Xiong, Caiming and Wu, Chien-Sheng},
  journal={arXiv preprint arXiv:2510.03663},
  year={2025}
}

@article{yu2025bbox,
  title={BBox DocVQA: A Large Scale Bounding Box Grounded Dataset for Enhancing Reasoning in Document Visual Question Answer},
  author={Yu, Wenhan and Chen, Wang and Qi, Guanqiang and Li, Weikang and Li, Yang and Sha, Lei and Xia, Deguo and Huang, Jizhou},
  journal={arXiv preprint arXiv:2511.15090},
  year={2025}
}

@article{bai2023qwen,
  title={Qwen-vl: A frontier large vision-language model with versatile abilities},
  author={Bai, Jinze and Bai, Shuai and Yang, Shusheng and Wang, Shijie and Tan, Sinan and Wang, Peng and Lin, Junyang and Zhou, Chang and Zhou, Jingren},
  journal={arXiv preprint arXiv:2308.12966},
  volume={1},
  number={2},
  pages={3},
  year={2023}
}

@inproceedings{chen2024internvl,
  title={Internvl: Scaling up vision foundation models and aligning for generic visual-linguistic tasks},
  author={Chen, Zhe and Wu, Jiannan and Wang, Wenhai and Su, Weijie and Chen, Guo and Xing, Sen and Zhong, Muyan and Zhang, Qinglong and Zhu, Xizhou and Lu, Lewei and others},
  booktitle={Proceedings of the IEEE/CVF conference on computer vision and pattern recognition},
  pages={24185--24198},
  year={2024}
}

@article{beyer2024paligemma,
  title={Paligemma: A versatile 3b vlm for transfer},
  author={Beyer, Lucas and Steiner, Andreas and Pinto, Andr{\'e} Susano and Kolesnikov, Alexander and Wang, Xiao and Salz, Daniel and Neumann, Maxim and Alabdulmohsin, Ibrahim and Tschannen, Michael and Bugliarello, Emanuele and others},
  journal={arXiv preprint arXiv:2407.07726},
  year={2024}
}

@article{zhang2023ideal,
  title={Ideal: Influence-driven selective annotations empower in-context learners in large language models. arXiv},
  author={Zhang, S and Xia, X and Wang, Z and Chen, LH and Liu, J and Wu, Q and Liu, T},
  journal={arXiv preprint arXiv:2310.10873},
  year={2023}
}

@inproceedings{wang2022cris,
  title={Cris: Clip-driven referring image segmentation},
  author={Wang, Zhaoqing and Lu, Yu and Li, Qiang and Tao, Xunqiang and Guo, Yandong and Gong, Mingming and Liu, Tongliang},
  booktitle={Proceedings of the IEEE/CVF conference on computer vision and pattern recognition},
  pages={11686--11695},
  year={2022}
}

@article{xiang2026safety,
  title={When Safety Collides: Resolving Multi-Category Harmful Conflicts in Text-to-Image Diffusion via Adaptive Safety Guidance},
  author={Xiang, Yongli and Hong, Ziming and Wang, Zhaoqing and Zhao, Xiangyu and Han, Bo and Liu, Tongliang},
  journal={arXiv preprint arXiv:2602.20880},
  year={2026}
}

@inproceedings{gao2024boosting,
  title={Boosting transferability in vision-language attacks via diversification along the intersection region of adversarial trajectory},
  author={Gao, Sensen and Jia, Xiaojun and Ren, Xuhong and Tsang, Ivor and Guo, Qing},
  booktitle={European Conference on Computer Vision},
  pages={442--460},
  year={2024},
  organization={Springer}
}

@article{jia2025semantic,
  title={Semantic-aligned adversarial evolution triangle for high-transferability vision-language attack},
  author={Jia, Xiaojun and Gao, Sensen and Guo, Qing and Qin, Simeng and Ma, Ke and Huang, Yihao and Liu, Yang and Tsang, Ivor and Cao, Xiaochun},
  journal={IEEE Transactions on Pattern Analysis and Machine Intelligence},
  year={2025},
  publisher={IEEE}
}

@article{jia2025adversarial,
  title={Adversarial attacks against closed-source mllms via feature optimal alignment},
  author={Jia, Xiaojun and Gao, Sensen and Qin, Simeng and Pang, Tianyu and Du, Chao and Huang, Yihao and Li, Xinfeng and Li, Yiming and Li, Bo and Liu, Yang},
  journal={arXiv preprint arXiv:2505.21494},
  year={2025}
}
